\documentclass[sigconf]{acmart}
\usepackage{enumitem}
\usepackage{multirow}
\usepackage{makecell}
\usepackage{algorithm}  
\usepackage{algorithmicx}  
\usepackage{algpseudocode}
\usepackage{color}
\usepackage{xcolor}
\usepackage{flushend}
\usepackage{stfloats}
\usepackage{colortbl}
\usepackage[normalem]{ulem}
\useunder{\uline}{\ul}{}

\newtheorem{theorem}{Theorem}

\newtheorem{definition}{Definition}

\AtBeginDocument{%
 }

\begin{document}

\author{
Zhengyu Wu,
Guang Zeng,
Huilin Lai,
Daohan Su, 
Jishuo Jia,
Yinlin Zhu,
Xunkai Li, \\
Rong-Hua Li, 
Guoren Wang, 
Chenghu Zhou
}

\title{Knowledge-Driven Federated Graph Learning on Model Heterogeneity}



\begin{abstract}
    Federated graph learning (FGL) has emerged as a promising paradigm for collaborative graph representation learning, enabling multiple parties to jointly train models while preserving data privacy.
    However, most existing approaches assume homogeneous client models and largely overlook the challenge of model-centric heterogeneous FGL (MHtFGL), which frequently arises in practice when organizations employ graph neural networks (GNNs) of different scales and architectures.
    Such architectural diversity not only undermines smooth server-side aggregation, which presupposes a unified representation space shared across clients’ updates, but also further complicates the transfer and integration of structural knowledge across clients. 
    To address this issue, we propose the Federated Graph Knowledge Collaboration (FedGKC) framework.
    FedGKC introduces a lightweight \textbf{Copilot Model} on each client to facilitate knowledge exchange while local architectures are heterogeneous across clients, and employs two complementary mechanisms: \textbf{Client-side Self-Mutual Knowledge Distillation}, which transfers effective knowledge between local and copilot models through bidirectional distillation with multi-view perturbation; and
    \textbf{Server-side Knowledge-Aware Model Aggregation}, which dynamically assigns aggregation weights based on knowledge provided by clients.
    Extensive experiments on eight benchmark datasets demonstrate that FedGKC achieves an average accuracy gain of 3.88\% over baselines in MHtFGL scenarios, while maintaining excellent performance in homogeneous settings.
\end{abstract}

\ccsdesc[500]{Computing methodologies~Semi-supervised learning settings}
\ccsdesc[500]{Computing methodologies~Neural networks}

\keywords{Federated Graph Learning; Heterogeneity; Knowledge Distillation}

\maketitle

\section{Introduction}
\label{sec: Introduction}
In today’s data-driven world, graphs have become indispensable for representing complex relationships in domains such as social networks~\cite{kumar2022socialgraph}, bioinformatics~\cite{zhang2021biograph}, and recommendations~\cite{he2020recomgraph}.
Graphs effectively capture relational attributes and provide rich insights for machine learning.
Graph Neural Networks (GNNs)~\cite{wu2020GNN} extend deep learning to non-Euclidean domains through message-passing mechanisms, enabling robust modeling that integrates local node features with global structural information.
However, the rising data volume and privacy concerns expose the limitations of centralized training, which risks data leakage and hinders cross-entity collaboration.
FGL~\cite{liu2024FGL} addresses these issues by training models locally and sharing only model updates, thus enabling collaborative learning while guaranteeing privacy and communication efficiency.

As FGL applications expand~\cite{long2020federatedProblem}, the disparity between clients in data volume, computational resources, and data quality becomes non-negligible.
Larger enterprises typically possess abundant data, allowing them to deploy large-scale GNNs~\cite{LargeCK, WebScale, jknet}.
In contrast, smaller companies with limited data may adopt lightweight or domain-specific models tailored to their unique operational contexts~\cite{velivckovic2018gat}.
Existing research in FGL has primarily addressed data-centric heterogeneity~\cite{ijcai2023FGSSL,li2023fedgta}, while paying limited attention to the more fundamental \textbf{model-centric heterogeneity challenge} (MHtFGL), where divergent model architectures across clients introduce incompatibilities in representation learning and hinder effective knowledge integration at the server.
Most FGL studies persuppose the homogeneous model architecture across clients, which curtails their applicability in practice.
To bridge this gap, we propose a novel FGL framework designed to tackle the MHtFGL challenge while enhancing performance of collaborative training.

From a model-centric perspective, the essence of FGL lies in the client–server paradigm, wherein locally learned implicit knowledge must be effectively communicated between clients and the server to facilitate collaborative learning.
The key to effective collaboration under MHtFGL lies in the choice of a proxy \textbf{Knowledge Carrier}, an alternative medium that replaces the shared model parameters~\cite{liang2020LG, yi2023fedgh}, which have long served as the default knowledge carrier but require an identical local model architecture for parameter alignment and convergence.
Prior works have proposed alternative knowledge carriers, which can be boardly summarized into three categories:  \textbf{(1) Prototype}~\cite{tan2022fedproto, zhang2024fedtgp, yi2023fedssa}: They condense locally learned node representations into class-wise prototype vectors, which are typically low-dimensional embeddings summarizing local semantics. 
\textbf{(2) Public/Synthetic Information}~\cite{fang2024noise, yang2023allosteric, zhang2024improving}: The server stores either the Public dataset or collaboratively trained synthetic Generators, which are utilized as the reference medium. They are distributed to local clients,  who then align their local training using knowledge carriers such as prediction logits, gradients, or generator parameters, etc.
\textbf{(3)Partial Model}~\cite{wu2024fiarse, alam2022fedrolex, lu2021heterogeneous}: The server either distributes a homogeneous model component shared across heterogeneous clients or allocates model subsets of varying depth or parameter scale according to each client’s computational capacity and the relevance of parameter subsets to its local data distribution. 
Within the context of FGL, \textbf{Parial Model} can be regarded as the most effective solution since the strength of GNNs stems from the neighbor aggregation and  propagation mechanism, through which implicit graph knowledge is tightly entangled with the trained model parameters themselves. In contrast, the knowledge carriers employed by the other two strategies rely primarily on model outputs (e.g., logits or gradients), which can only serve as indirect reflections of the underlying graph-structural knowledge.

\begin{figure}[t]
\centering
\includegraphics[width=\linewidth]{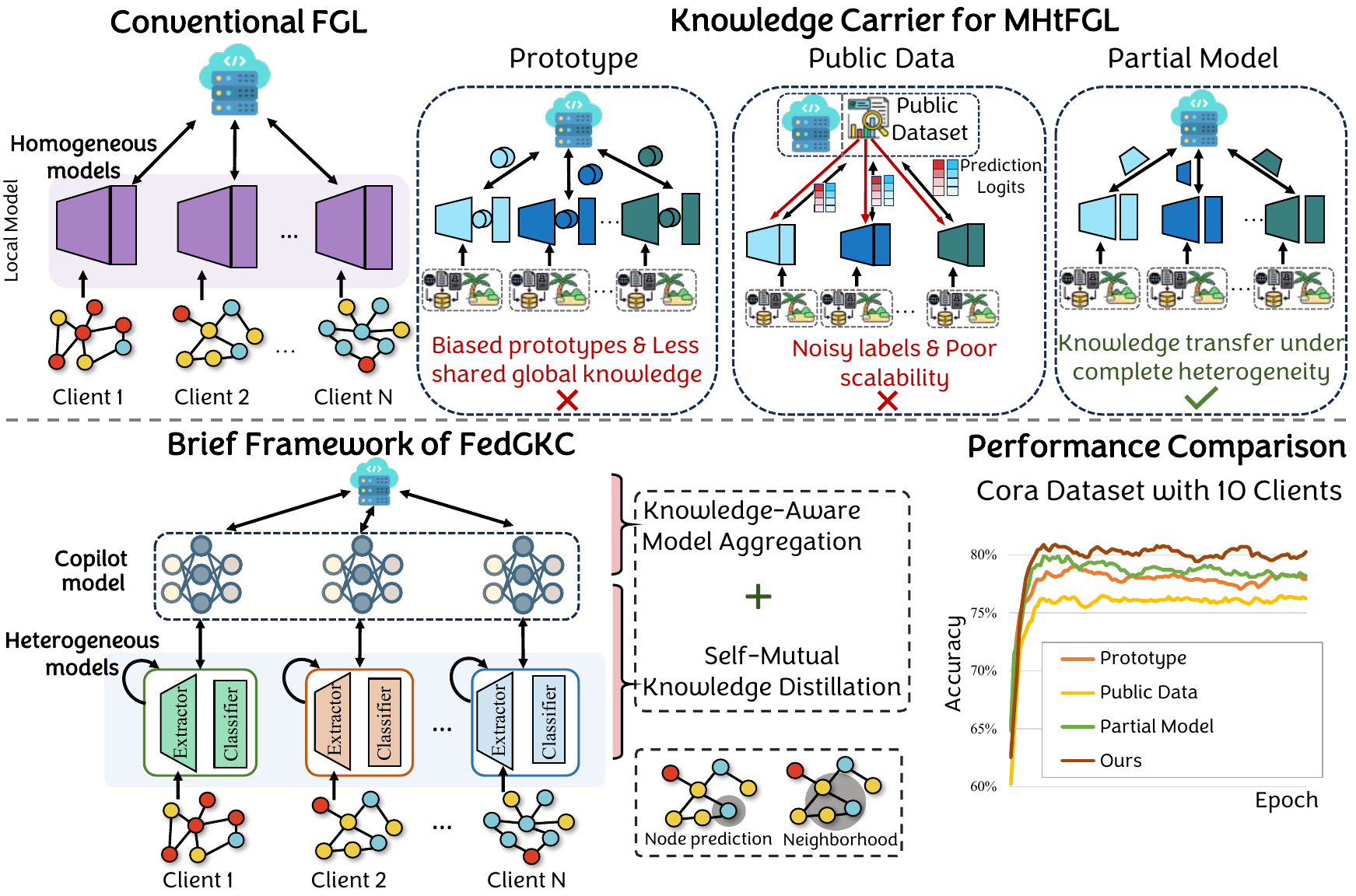}
\vspace{-6mm}
\caption{Illustration of the conventional FGL, heterogeneous federated learning approaches, and our proposed FedGKC.}
\vspace{-3mm}
\label{fig_intro}
\end{figure}

Although effective, existing Partial Model strategies face notable limitations.
Sharing activations or aligned model layers poses privacy risks and limits scalability as client numbers and diversity of local model architectures increase.
The exchanged submodels are typically topology-agnostic components (e.g., classification heads), whose aggregation yields suboptimal performance since graph knowledge is inherently topology-dependent.
To overcome these issues, we introduce a lightweight \textbf{Copilot Model} (illustrated in Fig.~\ref{fig_intro}) as a dedicated, architecture-agnostic knowledge carrier for cross-client transfer.
It acts as an auxiliary module trained via knowledge distillation coupled with topology-aware objectives, enabling it to capture both feature semantics and structural dependencies from the local training.
As shown in Fig.~\ref{fig_intro}, a toy experiment on the 10-client Cora dataset demonstrates that while Partial Model outperform other paradigms, our approach, as an innovative paradigm to Partial Model, achieves the best efficiency and effectiveness.
Implementation details are provided in Appendix~\ref{appendix:toy_example}.

Building on the above analysis, we propose the \textbf{Fed}erated \textbf{G}raph \textbf{K}nowledge \textbf{C}ollaboration (\textbf{FedGKC}) framework to address MHtFGL.
The framework leverages knowledge distillation while explicitly adapting it to structural properties of graph data.
To this end, we introduce copilot models on the client-side to bridge architectural differences and design a knowledge-aware aggregation strategy on the server-side to ensure effective integration of diverse knowledge uploaded from clients.
Specifically, FedGKC consists of two key components:
\textbf{(1) Self-Mutual Knowledge Distillation (SMKD)} on heterogeneous clients.
We establish a bidirectional knowledge distillation process between each local odel and its copilot model.
This design enables the copilot to acquire localized knowledge while simultaneously injecting global knowledge back into the local model.
Beyond class-level predictions, SMKD also transfers neighborhood information, thereby enhancing the model’s ability to capture topological dependencies. (Sec.~\ref{subsec: mutual distillation})
To improve robustness, we incorporate a multi-view data perturbation strategy for local self-distillation , ensuring consistent predictions across perturbed local graphs and boosting generalizabilities. (Sec.~\ref{subsec: self distillation})
\textbf{(2) Knowledge-Aware Model Aggregation (KAMA)} on the server side.
Unlike the conventional FedAvg algorithm~\cite{2017FedAvg}, which assigns weights to clients solely based on their data volume,  KAMA recognizes that heterogeneous clients differ significantly in learning capacity and knowledge quality.
Specifically, we design an adaptive weighting mechanism based on data volume, knowledge strength, and knowledge clarity.
Performing aggregation on knowledge-aware weights, KAMA achieves more accurate and efficient integration between clients, guiding the global model toward better convergence.

To summarize, our work offers three following contributions:
\textbf{(1) Problem Identification.}
We are the first to systematically investigate the challenge of MHtFGL, specifically the partial model strategy, where diverse client architectures hinder unified training and complicate the integration of graph knowledge in real-world deployments.
\textbf{(2) Innovative Framework.}
We propose \textbf{FedGKC}, a knowledge-driven FGL framework that enables effective collaboration across clients with heterogeneous models.
It incorporates two key mechanisms: SMKD, which bridges structural differences through bidirectional and topology-aware knowledge distillation between local and copilot models; and KAMA, which adaptively aggregates client contributions based on knowledge quality, ensuring more reliable global insights.
\textbf{(3) Superior Performance.} We validate FedGKC on eight benchmark datasets under both heterogeneous and homogeneous settings. Experiments show consistent improvements, with an average accuracy gain of 3.88\% and up to 8.13\% over state-of-the-art baselines, confirming both the effectiveness and generalizability of our approach.

\vspace{-4mm}
\section{Related Works}
\label{sec: Related Works}

\subsection{Federated Graph Learning}
\label{sec: Federated Graph Learning}
    FGL extends FL to GNNs and enables collaborative training on graph-structured data while preserving data privacy.
    Instead of exchanging raw data, clients upload learned graph representation models to the secured server to aggregate cross-client knowledge.
    Existing FGL research can be broadly categorized into two settings:
    \textbf{(1) Graph-FL}: Each client owns multiple independent graphs and the goal is to address graph-level tasks (e.g., graph classification).
    Representative approaches include GCFL~\cite{xie2021GCFL} and FedStar~\cite{tan2023fedstar},etc.    
    \textbf{(2) Subgraph-FL}: Each client holds a subgraph of an implicit global graph and the objective is to solve node-level tasks (e.g., node classification).
    Representative works include FGSSL~\cite{ijcai2023FGSSL}, FedGTA~\cite{li2023fedgta}, FedPub~\cite{baek2023fedpub}, and FedSage+~\cite{zhang2021fedsage}.
    Notably, previous studies have concentrated on addressing data heterogeneity challenges ~\cite{ijcai2023FGSSL,li2023fedgta}.
    However, in real-world scenarios, GNNs from different clients may deploy divergent architectural models, posing fundamental obstacles to unified graph learning and complicating collaborative training.
    To the best of our knowledge, we are the first to investigate this issue and propose the new paradigm tailored for this challenge.

\subsection{Model-Centric Heterogeneous FL}
\label{sec: Model-Centric Heterogeneous Federated Learning}
    Model-centric heterogeneous federated learning (MHtFL) enables flexible and diverse client architectures while preserving data privacy. Existing MHtFL methods can be broadly classified into three categories, each with distinct limitations: \textbf{(1) Prototype}~\cite{tan2022fedproto, zhang2024fedtgp, tan2022fedpcl} avoid sharing parameters but suffer from biased classifiers, further exacerbated by architectural diversity~\cite{li2023bias}, which undermines efficacy of the global knowledge aggregation. 
    \textbf{(2) Public/Synthetic Information}~\cite{fang2024noise, yang2023allosteric, zhang2024improving, zhang2021parameterized, lin2020ensemble} distribute globally stored public dataset or generators for syntethizing graph information to clients. During local training, heterogeneous clients align their predictions or gradients on shared public data or upload generator parameters for aggregation. However, such proxies act as indirect knowledge carriers and fail to capture the topology-dependent structural knowledge that is crucial for effective FGL training. 
    \textbf{(3) Partial Model} methods such as LG-FedAvg~\cite{liang2020LG}, FedGen~\cite{zhu2021fedgen}, and FedGH~\cite{yi2023fedgh} permit heterogeneity in major model components while enforcing homogeneity on a small shared module. This design enables limited global knowledge transfer but constrains model expressiveness. Approaches such as FML~\cite{shen2023fedfml} and FedKD~\cite{wu2022communication} employ auxiliary global models for mutual distillation; however, early training rounds risk exchanging low-quality or unstable knowledge.

    Notably, these approaches are not specifically tailored for graph-structured data. To address the issue, we propose an approach that employs knowledge distillation to facilitate effective graph information transfer between local heterogeneous models and a globally shared Copilot Model during local training, thereby balancing personalized local knowledge and global generalizability.

\subsection{Knowledge Distillation on Graph}
\label{sec: Knowledge Distillation on Graph}
    Knowledge distillation (KD)~\cite{hinton2015distilling} transfers knowledge from large models to smaller ones, reducing computational cost while preserving performance.
    In graph learning, KD enables knowledge transfer across diverse models, with the main challenge being what knowledge to distill and how to distill it.
    
    Three main categories of transferable knowledge are commonly explored: 
    \textbf{(1) Logits}, i.e., outputs before the Softmax layer, where methods minimize distribution divergence to transfer predictive knowledge~\cite{wang2021MulDElogitsKD1,yan2020logitsKD2,yang2021logitsKD3}.
    \textbf{(2) Graph structure}, which captures node-edge connectivity and relationships, with approaches aligning nodes via neighborhood relations to transfer structural information~\cite{yang2020structKD1, guo2022structKD2}.
    \textbf{(3) Embeddings}, i.e., intermediate node representations extracted from hidden layers, which contain feature and structural context. It can be distilled to guide another model’s training~\cite{huo2023embeddingKD1,he2022embeddingKD2,yu2022embeddingKD3}.
    
    Regarding distillation methods, direct distillation minimizes differences between models, promoting intuitive knowledge transfer~\cite{yan2020logitsKD2}; adaptive distillation offers flexibility by considering the importance of different knowledge components.
    Methods like RDD~\cite{zhang2020RDD}, FreeKD~\cite{feng2022freekd}, and MulDE~\cite{wang2021MulDElogitsKD1} adjust the distillation process based on the certainty or correctness of the knowledge, allowing models to focus more on reliable information while avoiding knowledge that is less informative or noisy.
    Furthermore, various distillation strategies tailored for specific tasks exist within machine learning~\cite{he2022embeddingKD2}.
    Our study aims to ensure the preservation and transfer of effective graph information within heterogeneous federated scenarios, thereby facilitating collaborative training among multiple clients.

\section{Methodology}
\label{sec: Methodology}

\begin{figure*}[ht]
\centering
\includegraphics[width=0.9\linewidth]{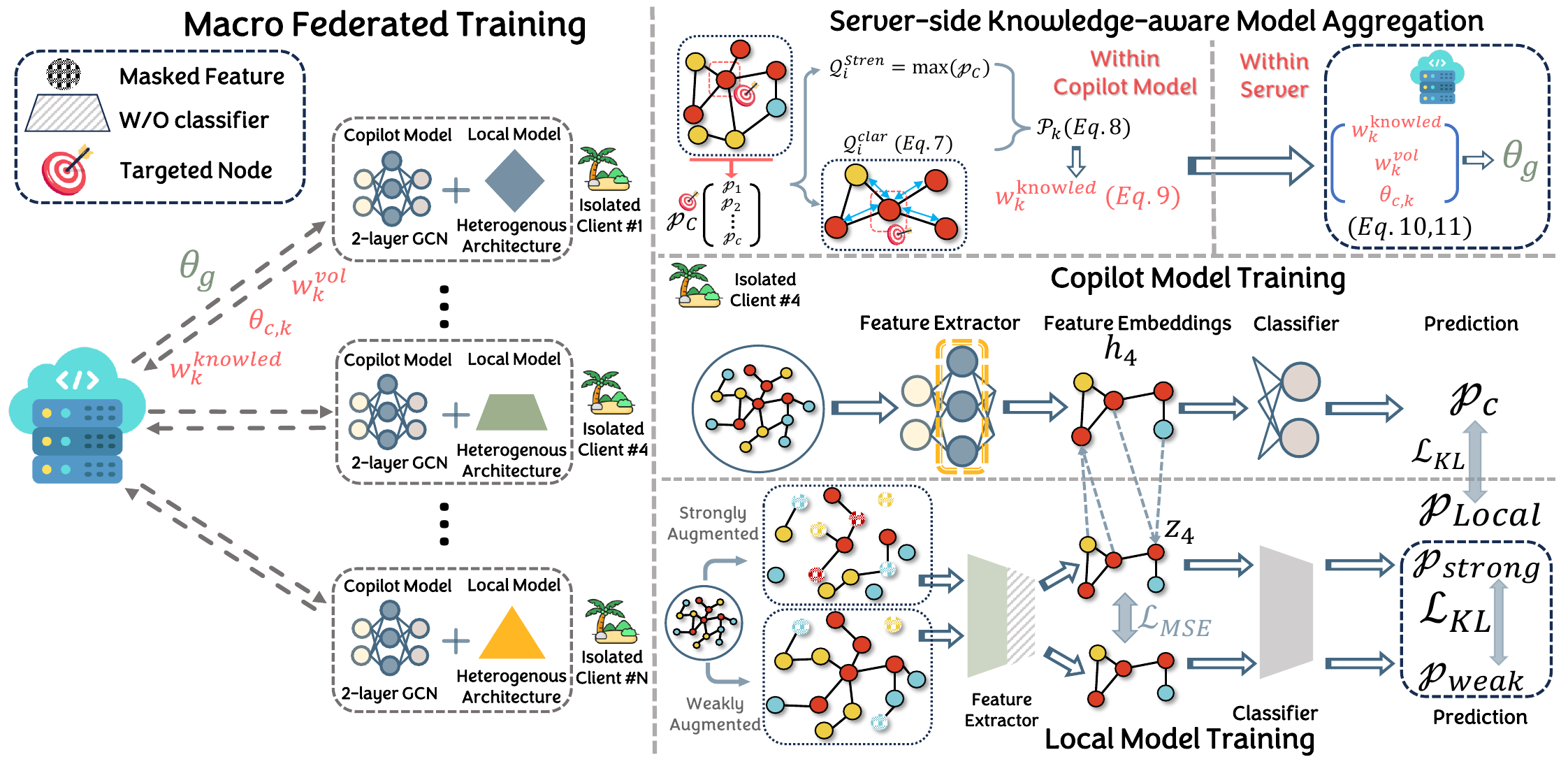}
\vspace{-3mm}
\caption{The visualized training pipeline of FedGKC. The Macro Federated Training depicts the FGL architecture in its essence, in which each client uploads its data volume: $w_k^{vol}$,  trained copilot model parameters: $\theta_{c,k}$, and knowledge-realted weights: $w_k^{knowled}$, to secured server for model aggregation. On clients, heterogenous model architectures are implemented, which are visually separated by different shapes and colors. The right-hand side of the figure use the Isolated Client \#4 showcases the training. Notably, to have a better visual illustration for readers, we use fewer nodes from the same subgraph dataset to increase its display clarity. All symbols in the figure are aligned with our methodological descriptions in Sec.~\ref{sec: Overview}.  Copilot Model Training and Local Model Training refer to Sec.~\ref{subsec: mutual distillation}, and Server-side Knowledge-aware Model Aggregation refers to Sec.~\ref{sec: Knowledge-Aware Model Aggregation}. }
\vspace{-2mm}
\label{fig_method}
\end{figure*}

\subsection{Notations and Problem Formulation}
\label{sec: Notations and Problem Formulation}
    In this study, we propose an MHtFGL framework designed to leverage diverse graph data and GNN models from multiple clients for collaborative training, while ensuring the protection of data privacy. 

    Let there be $K$ clients, where the graph data owned by client $k$ is represented as $G_k = (V_k, E_k)$ with $|V_k| = N_k$ nodes and $|E_k| = M_k$ edges.
    The graph's adjacency matrix, node feature matrix, and label matrix are denoted as $A_k \in \mathbb{R}^{N_k \times N_k}$, $X_k \in \mathbb{R}^{N_k \times f}$, and $Y_k \in \mathbb{R}^{N_k \times C}$, respectively, where $f$ represents the number of node features and $C$ denotes the number of classes.
    Note that the graph data and the GNN model architectures can vary among different clients.
    
    GNNs constitute a class of deep learning models tailored to perform embedding and inference tasks on graph-structured data.
    Federated learning is a collaborative learning paradigm that enables the training of a shared model across multiple data owners without the direct exchange of raw data.
    Each training round involves the selection of participating clients, the update of the local model via local training on each chosen client, the aggregation of local model parameters on a central server, and the subsequent distribution of the aggregated parameters back to the clients.
    The objective of our work is to enhance the performance of each client’s local model on node classification tasks in heterogeneous environments.

\subsection{Overview}
\label{sec: Overview}
    In this section, we properly present our proposed FedGKC, whose overall training architecture is illustrated in Fig.~\ref{fig_method}.
    At first, the parameters of all models are initialized.
    Then, the following steps are executed in each round of federated training:
    (1) Local model and its received copilot model on each client are trained following objectives defined in Self-Mutual Knowledge Distillation (SMKD).
    (2) At the server, FedGKC performs
    Knowledge-Aware Model Aggregation (KAMA) with uploaded messages indicating local knowledge strength and clarity alongside updated copilot model parameters from each client.
    (3) The aggregated model parameters are distributed back to all client copilot models.
    Steps (1) to (3) are repeated until convergence, allowing each client's local model to learn global knowledge without transmitting graph data, guaranteeing on alignning the local and global optimizing direction.

\subsection{Self-Mutual Knowledge Distillation}
    The model learning process on the client side consists of two main components.
    Firstly, there is a mutual knowledge learning process between the local model and the copilot model.
    This process not only allows the copilot model to acquire more localized knowledge but also enables the local model to obtain global knowledge that includes information from other clients.
    Secondly, to enhance the predictive capability of the local model, we generate two different views by perturbating graph elements from the same graph to implement self-distillation.
    This ensures that local model generates consistent and robust embeddings and predictions.

\subsubsection{Mutual distillation addressing model heterogeneity}
\label{subsec: mutual distillation}
    In traditional knowledge distillation methods~\cite{yang2021logitsKD3}, knowledge is typically transferred by aligning the prediction results of different models on the same data.
    However, merely aligning the prediction results may limit the knowledge learned by the models, resulting in a lack of comprehensiveness and representativeness.
    In the context of graph data processed by GNNs, node representations are primarily derived from capturing neighborhood information.
    Therefore, we propose propagating neighborhood knowledge to enhance the utilization of topological information captured by GNNs:
\begin{equation}
    \mathcal{L}_{\mathrm{neigh}}=\frac{1}{|\mathcal{N}|} \sum_{i \in \mathcal{N}} \sum_{j \in \mathcal{V}_i \cup i} \mathcal{D}\left(\sigma\left({z}_j\right), \sigma\left({h}_i\right)\right),
\end{equation}
    where $z$ and $h$ represent the embedding outputs of the feature extractor of the two models, respectively.
    $\mathcal{V}_i$ represents the set of neighbor nodes of node $i$, $\mathcal{N}$ denotes the total number of nodes, $\sigma= softmax(\cdot)$ denotes an activation function, and $ \mathcal{D}(\cdot)$ denotes a distance function (e.g., Kullback Leibler divergence~\cite{hinton2015distilling}).
    
    For the copilot model, the optimization objective is denoted as: 
\begin{equation}
\label{equ:copmodel}
    \mathcal{L}_{\text{cop-mutu}} = \alpha \mathcal{L}_{CE}\left(y, p_c\right) + \beta \mathcal{L}_{\mathrm{neigh}} + \left(1-\alpha-\beta\right) \mathcal{L}_{KL}\left(p_l\Vert p_c\right),
\end{equation}
    where $\alpha$ and $\beta$ are the weights to balance the influence of different distillation losses, and $p_l$ and $p_c$ are the predictions of local and copilot models, respectively.
    Here, $\mathcal{L}_{KL}$ is the KL divergence function, and $\mathcal{L}_{CE}\left(y, p_c\right)=-y\log\left(softmax\left(p_c\right)\right)$ represents the cross-entropy loss between $p_c$ and the truth label $y$.

    Similarly, the local model learn information from the copilot model and train on local data, and its optimization objective is:
\begin{equation}
\label{equ:localmutu}
    \mathcal{L}_{\text{local-mutu}} = \alpha \mathcal{L}_{CE}\left(y, p_l\right) + \beta \mathcal{L}_{\mathrm{neigh}} + \left(1-\alpha-\beta\right) \mathcal{L}_{KL}\left(p_c\Vert p_l\right).
\end{equation}

\subsubsection{Self-distillation mining model potential}
\label{subsec: self distillation}
    To further stimulate the performance of the local model, we introduce a self-distillation strategy.
    We first create two perspectives on the same graph data, a strongly perturbated view and a weakly perturbated view.
    This dual-perspective strategy effectively enhances the diversity of graph data by applying varying degrees of edge removal and feature masking~\cite{zhao2022graphAug}.
    The weakly perturbated view retains more original information, while the strongly perturbated view significantly obscures and removes more graph information (e.g., topological structure, node semantics).
    We expect the predictions of the strongly perturbated view to closely align with the predictions of the weakly perturbated view for the same input.
    It aims to ensure that, even when faced with substantial information alteration, the local model maintains consistency in its predictions, thereby deepening its understanding of the data's intrinsic nature.
    The optimization objective for the self-distillation of the local model is denoted as 
\begin{equation}
    \mathcal{L}_{\text{self-distill}} = \mathcal{L}_{MSE}\left(e_l^{\mathrm{weak}}, e_l^{\mathrm{strong}}\right) + \mathcal{L}_{KL}\left(p_l^{\mathrm{weak}}\Vert p_l^{\mathrm{stong}}\right),
\end{equation}
    where $\mathcal{L}_{MSE}$ denotes the mean-square error loss, and $e$ and $p$ represent the embedding outputs of the model feature extractor and the prediction outputs of the classifier, respectively.
    In summary, the overall optimization objective for the local model integrates the mutual distillation objective with the self-distillation objective:
\begin{equation}
\label{equ:local}
    \mathcal{L}_{\text{local}} = \mathcal{L}_{\text{local-mutu}}+\mathcal{L}_{\text{self-distill}}.
\end{equation}
    The optimization goal for the copilot model is shown in Eq.~(\ref{equ:copmodel}).
    This coordinated optimization enhances the learning efficiency and knowledge acquisition capability of the local model in MHtFGL.
 
\subsection{Knowledge-Aware Model Aggregation}
\label{sec: Knowledge-Aware Model Aggregation}
    Given the heterogeneity among clients, the learning efficiency of each client's model varies accordingly.
    Traditional FL methods typically employ an averaging aggregation strategy~\cite{2017FedAvg}, which does not take into account the model heterogeneity scenario.
    This oversight results in conflicted models aggregation and suboptimal overall performance.
    To address it, we propose a Knowledge-Aware Model Aggregation mechanism that dynamically assigns weights based on learned knowledge quality of each client's model.
    Through dynamic weighting, well-informed clients is protected from the adverse effects of clients with limited knowledge.
    Simultaneously, it channels greater guidance from knowledgeable clients to less-informed ones, improving convergence and robustness.

    Specifically, when aggregating the copilot models from each client, the server first considers the \textbf{data volume} from an overall perspective and calculates volume weight accordingly, denoted as: 
\begin{equation}
    w_k^{vol}=\frac{\mathcal{N}_k}{\sum_{k=1}^K \mathcal{N}_k},
\end{equation}
    where $\mathcal{N}_k$ indicates the amount of data on the $k$-th client. 

    Next, it assesses the knowledge levels learned by each node in the copilot model.
    Ideally, each node should focus on relevant knowledge associated with its respective category, implying that the knowledge should be strong and clear.
    We measure knowledge level from two dimensions: \textbf{knowledge strength} and \textbf{knowledge clarity}.
    The knowledge strength at a node can be represented by the maximum value of the output prediction probabilities, which intuitively reflects the model's confidence in recognizing a particular category, denoted as $Q_i^{stren} = \max\left(p_i\right)$.
    On the other hand, assessing knowledge clarity is more complex as it involves evaluating the distinction between the predictions of this node and those of other categories, expressed as $Q_i^{stren}-\sum{p_i^*}$, where $p^*$ represents the set of values other than the maximum value.
    Additionally, we must be wary of the over-smoothing phenomenon in GNNs due to interactions with neighbor nodes, which compromises the model's discriminatory ability.
    The clarity evaluation is formulated as:
\begin{equation}
\label{equ:clarity}
    Q_i^{clar} = \frac{1}{M-1}\left(Q_i^{stren}-\sum{p_i^*}\right) - \lambda\left(\frac{1}{|\mathcal{V}_i|}\sum_{j \in \mathcal{V}_i} \text{sim}\left(p_i, p_j\right)\right),
\end{equation}
    where $M$ represents the number of categories, $\mathcal{V}_i$ represents the set of neighbor nodes of node $i$, $\lambda$ is the weight of the over-smoothing effect, and $\text{sim}(\cdot)$ represents the cosine similarity calculation.
    Then, the degree of knowledge mastery of each client is quantified by:
\begin{equation}
\label{equ:know}
    \mathcal{P}_k=\frac{1}{\mathcal{N}_k}\sum_{i=1}^{\mathcal{N}_k} \left(Q_i^{stren}+Q_i^{clar}\right),
\end{equation}
    where $\mathcal{N}_k$ indicates the amount of data on the $k$-th client.
    The knowledge-related weights are expressed as:
\begin{equation}
    w_k^{knowled}=\frac{\mathcal{P}_k}{\sum_{k=1}^K \mathcal{P}_k}.
\end{equation}

    By considering data volume, node knowledge strength, and node knowledge clarity, we quantify the learning knowledge level of each client's copilot model.
    Consequently, the aggregation weight for the $k$-th client's model can be determined as follows:
\begin{equation}
\label{equ:weighttotal}
w_k=\left(w_k^{vol}+w_k^{knowled}\right)/2.
\end{equation}
    The model aggregation formula is:
\begin{equation}
\label{equ:genera}
\theta_g=\sum_{k=1}^K w_k \cdot \theta_{c,k},
\end{equation}
    where $\theta_{c,k}$ is the parameters of the $k$-th client copilot model.

    Finally, the aggregated model parameters will be distributed back to each client's copilot model in preparation for the next round of client training.
    Through this meticulous adjustment, our aggregation strategy effectively integrates multi-source information, achieving a more efficient and equitable model aggregation process.
    The complete algorithm of FedGKC is presented in Algorithm~\ref{algorithm}.

\begin{algorithm}[t]
\caption{The complete algorithm of FedGKC}
\label{algorithm}
\begin{algorithmic}[1]
\Require
Communication round $T$; Initial model parameters $\theta$; The number of clients $K$;
\Ensure
Local models for all clients;
\State Initialize all client-side copilot and local models;
\For{$t=1,\cdots, T$}
\State $//$ \textbf{Client Side}:
\For{$k=1,\cdots, K$ in parallel}
\State Optimize the coplit model according to Eq.~(\ref{equ:copmodel});
\State Optimize the local model according to Eq.~(\ref{equ:local});
\State Record the data volume $\mathcal{N}_k$;
\State Calculate the knowledge level $\mathcal{P}_k$ through Eq.~(\ref{equ:know});
\EndFor
\State $//$ \textbf{Server Side}:
\State Compute the model aggregation weights ${w}_k$ by Eq.~(\ref{equ:weighttotal});
\State Update aggregate model parameters by Eq.~(\ref{equ:genera});
\State Send aggregate model parameters to the client-side copilot model: $\theta_{c,k} \leftarrow \theta_{g}$;
\EndFor
\end{algorithmic}
\end{algorithm}

\subsection{Complexity Analysis}
\label{sec: Complexity Analysis}
    We now provide a computational complexity analysis of FedGKC.
    Each client employs a GNN as the local model.
    For an $l$-layer GNN model with a batch size $b$, the propagated feature $X^{(l)}$ has a space complexity of $\mathcal{O}((b+l)f)$, and the overhead for linear regression operations is $\mathcal{O}(f^2)$, where $f$ are the number of feature dimensions.
    The introduction of the copilot model doubles the parameter storage space, leading to an overall space complexity of $\mathcal{O}(2(b+l)f+2f^2)$.
    Furthermore, the inclusion of additional alignment losses within the self-mutual knowledge distillation module introduces a time complexity of $\mathcal{O}(f^2)$.
    For the server, the reception and aggregation of model parameters impose space and time complexities of $\mathcal{O}(Nf^2)$ and $\mathcal{O}(N)$, respectively, where $N$ represents the number of clients involved.
    Therefore, the total space complexity is dominated by client-side storage and server-side aggregation, yielding $\mathcal{O}(2N(b+l)f+2Nf^2)$.
    The total time complexity per communication round is primarily governed by client-side local training and server-weighted aggregation, resulting in $\mathcal{O}(N(b+l)f+Nf^2)$.
    Compared to conventional FGL methods, FedGKC introduces manageable overhead through its knowledge distillation and knowledge-aware aggregation mechanisms, achieving an effective balance between efficiency and performance.
    The detailed analysis and comparsion can be found in Appendix~\ref{app:complexity}.

\section{Theoretical Analysis}
\label{sec: Theoretical Analysis}
    To substantiate the effectiveness of FedGKC and improve its theoretical interpretability, we show that SMKD enhances stability and robustness, while KAMA drives the descent direction closer to the global optimum compared to conventional federated learning.
    
\subsection{Stability and Robustness of SMKD}
    To rigorously characterize the stability of embeddings under perturbation, we introduce a finite difference Lipschitz constant.
    In conjunction with the definition of our self-distillation loss and its associated inequality relations, we have Definition~\ref{Lipschitz Constant} and Theorem~\ref{Lipschitz Constant Equation}.
\begin{definition}[Lipschitz Constant]
\label{Lipschitz Constant}
    For each sample $i$, we define the perturbation gap between strong and weak views as $\Delta_i$ and according to the perturbation gap, we define the finite difference Lipschitz constant of embeddings as:
\begin{equation}
\label{Lipschitz Constant Definition}
    \Delta_i := x_i^{\text{strong}} - x_i^{\text{weak}}, \;\;\; \hat{L}_e(i) := \frac{\| e(x_i^{\text{strong}}) - e(x_i^{\text{weak}}) \|}{\| \Delta_i \|}.
\end{equation}
\end{definition}

\begin{theorem}
\label{Lipschitz Constant Equation}
    The squared discrepancy between the strong and weak perturbations, measured by the normalized embedding difference, is upper bounded by the ratio of the self-distillation loss to the average squared perturbation magnitude.
\begin{equation}
\label{Lipschitz Constant Inequality}
    \frac{1}{N} \sum_{i=1}^{N} \hat{L}_e(i)^2 \leq \frac{L_{\text{self}}}{\frac{1}{N} \sum_{i=1}^{N} \|\Delta_i\|^2}.
\end{equation}
\end{theorem}
    As training progresses, $L_{\text{self}}$ gradually decreases, leading to a corresponding reduction of the Lipschitz constant. Consequently, it demonstrates that the embeddings become insensitive to perturbations in the input data.

\subsection{Gradient Direction in KAMA}
    We first define the local descent direction of the $k$-th client as $g_k$, the ideal optimal descent direction as $g^*$, and the alignment measure as $a_k = \langle g_k, g^\ast \rangle$. According to the definition of knowledge clarity and our formulation of the aggregation weights, we obtain Theorem~\ref{Node Score Expression}.
\begin{theorem}
\label{Node Score Expression}
    The node score $S_i$ is defined as the sum of knowledge strength and knowledge clarity, where $\overline{\mathrm{sim}}_i$ represents the mean similarity among neighboring nodes of node i.
\begin{equation}
\label{Node Score}
    S_i:=Q_i^{\text{stren}}+Q_i^{\text{clar}}= \left(1+\frac{2}{M-1}\right) m_i-\frac{1}{M-1}-\lambda\,\overline{\mathrm{sim}}_i.
\end{equation}
\end{theorem}
    Furthermore, it can be derived that the node score $S_i$ exhibits monotonic growth with respect to the highest probability $m_i$ and monotonic decline with respect to the average neighborhood similarity. Therefore, our aggregation weights effectively capture greater knowledge quality. The magnitude of the aggregation weight is positively correlated with the degree of alignment.

    Based on our definition, the inner products of the two aggregated directions with $g_i$ are given as follows:
\begin{equation}
\label{Inner product Definition}
    \langle \hat{g}_{\mathrm{FedAvg}}, g^\star \rangle = \sum_k w_k^{\mathrm{vol}} a_k, \;\;\; \langle \hat{g}_{\mathrm{KAMA}}, g^\star \rangle = \sum_k w_k a_k.
\end{equation}
    Thus, we have our Theorem~\ref{Inner Product differences}.
\begin{theorem}
\label{Inner Product differences}
    The difference between the inner products of the two aggregated directions can be derived as the following expression, where $\overline{\mathrm{P}}$ and $\overline{\mathrm{a}}$ denote the average aggregation weight and alignment across clients, respectively.
{\small\begin{equation}
\label{Differences Equation}
    \langle\hat{g}_{\mathrm{KAMA}}, g^\star \rangle - \langle \hat{g}_{\mathrm{FedAvg}}, g^\star \rangle= \frac{1}{2}\left(\bar{a} - \sum_k w_k^{\mathrm{vol}} a_k+\frac{\mathrm{Cov}(P,a)}{\bar{P}}\right).
\end{equation}}
\end{theorem}
    Under the unbiased sampling condition, the first term on the right-hand side can be regarded as zero.
    Moreover, by Theorem~\ref{Node Score Expression}, $\mathrm{Cov}(P,a)$ is positive.
    Consequently, it demonstrates that the KAMA module ensures the descent direction of optimization is closer to the global optimum compared to conventional FL approaches such as FedAvg.
    Detailed proofs of Theorems are in the Appendix~\ref{app: proof}.

\section{Experiments}
\label{sec: Experiments}
    In this section, a comprehensive evaluation of FedGKC is provided.
    We aim to answer following key questions:
    \textbf{Q1}: Does FedGKC exhibit superior performance compared to state-of-the-art baseline methods?
    \textbf{Q2}: What contributes to the performance enhancement of FedGKC?
    \textbf{Q3}: Is FedGKC capable of generalizing effectively across varying copilot model architectures?
    \textbf{Q4}: How resilient is the performance of FedGKC to fluctuations in hyperparameter values?

\subsection{Experimental Setup}
\label{sec: Experimental Setup}

\subsubsection{Datasets}
\label{sec: Datasets}
     Experiments are conducted utilizing 8 commonly employed benchmark datasets in graph learning.
     These consist of three small-scale citation networks (Cora, CiteSeer, PubMed)~\cite{CoraDataset}, two medium-scale co-authorship networks (CS, Physics)~\cite{CSandPhyDataset}, two medium-scale user-item datasets (Photo, Computer)~\cite{mcauley2015ComputDataset}, and one large-scale open graph benchmark dataset (ogbn-arxiv)~\cite{ogbnarxiv}.
     Further particulars regarding these datasets are delineated in Table~\ref{tab:dataset}.

\subsubsection{Baselines}
    We compare FedGKC with heterogeneous federation strategies.
    Given the absence of strategies for MHtFGL, we re-implemented these methods under FGL conditions based on the available source code.
    There are personalization-based strategies (e.g., LG-FedAvg~\cite{liang2020LG}), prototype-based approaches (e.g., FedProto~\cite{tan2022fedproto}, FedTGP~\cite{zhang2024fedtgp}), knowledge distillation-based methods (e.g., FML~\cite{shen2023fedfml}, FedKD~\cite{wu2022communication}), partial-model based model FIARSE~\cite{wu2024fiarse} and method for MHtFGL called FedPPN~\cite{liu2024model}.
    Due to the need for partial homogeneity of the personalization method, we set up the same classifier for all clients when implementing the LG-FedAvg.
    In addition, we maintain a completely heterogeneous framework for all client models.
    The detailed implementations are in Appendix~\ref{app: Detailed Implementation of Experiment}.

\subsubsection{Heterogeneity Simulations}
\label{subsubsec:heterogeneity simulations}
    To partition data to each client,
    we deploy Louvain~\cite{Louvain} community detection algorithm to partition graph into subgraphs that are distributed to clients.
    To ensure uniformity and simulate multi-clients scenarios, we conduct experiments using configurations of 5, 10, and 20 clients.
    Specifically, we consider two model heterogeneity scenarios:
    \textbf{(1) Divergent model architectures}. We utilize five distinct two-layer graph neural network architectures: GCN~\cite{kipf2017GCN}, GAT~\cite{velivckovic2018gat}, GraphSAGE~\cite{hamilton2017graphsage}, GIN~\cite{xu2018gin}, and SGC~\cite{wu2019sgc}.
    \textbf{(2)Varying model scales}.
    We leverage deep GNNs technology ~\cite{jknet} to employ a lightweight two-layer SGC, a conventional two-layer GCN, and deeper models including four-layer GCN, six-layer GCN, and eight-layer GCN.
    For the allocation of clients, given $K$ clients, the model architecture assigned to the $k$-th client is determined by ($k \bmod 5$), $k \in |K|$.
    The parameters of each assigned model architecture are re-initialized for every client.

\begin{table*}[!t]
\caption{Performance comparison in \textcolor{magenta}{Divergent Model Architectures}. The best and sub-best results are marked in \textbf{\textcolor{red}{Red}} and \textbf{\textcolor{blue}{Blue}}.}
\vspace{-3mm}
\label{tab:performance}
\resizebox{\linewidth}{!}{
\begin{tabular}{lcccccccccccc}
\Xhline{1pt}
\rowcolor{gray!20} & \multicolumn{3}{c}{Cora} & \multicolumn{3}{c}{CiteSeer} & \multicolumn{3}{c}{PubMed} & \multicolumn{3}{c}{ogbn-arxiv} \\
\cline{2-13} \rowcolor{gray!20} \multirow{-2}{*}{Method} & 5 & 10 & 20 & 5 & 10 & 20 & 5 & 10 & 20 & 5 & 10 & 20 \\
\hline
LG-FedAvg & 78.78$_{\pm0.33}$ & 78.11$_{\pm0.12}$ & 70.52$_{\pm0.58}$ & \textbf{\textcolor{blue}{65.64$_{\pm0.47}$}} & 62.74$_{\pm0.29}$ & 58.57$_{\pm0.64}$ & 84.52$_{\pm0.36}$ & 82.74$_{\pm0.52}$ & 80.95$_{\pm0.27}$ & 61.32$_{\pm0.41}$ & 61.29$_{\pm0.38}$ & 61.62$_{\pm0.49}$ \\
FedProto & 78.50$_{\pm0.77}$ & 77.75$_{\pm0.22}$ & 74.13$_{\pm0.27}$ & 59.77$_{\pm1.17}$ & 59.26$_{\pm0.77}$ & 57.28$_{\pm0.81}$ & 84.71$_{\pm0.25}$ & 82.77$_{\pm0.04}$ & 81.74$_{\pm0.13}$ & 38.26$_{\pm2.13}$ & 44.82$_{\pm1.92}$ & 48.19$_{\pm0.73}$ \\
FML & \textbf{\textcolor{blue}{80.33$_{\pm0.44}$}} & \textbf{\textcolor{blue}{79.90$_{\pm0.39}$}} & \textbf{\textcolor{blue}{75.37$_{\pm0.52}$}} & 64.67$_{\pm0.46}$ & 62.52$_{\pm0.41}$ & 60.39$_{\pm0.57}$ & 84.39$_{\pm0.33}$ & 83.05$_{\pm0.48}$ & 82.23$_{\pm0.36}$ & 52.15$_{\pm0.71}$ & 58.25$_{\pm0.65}$ & 60.98$_{\pm0.49}$ \\
FedKD & 80.32$_{\pm1.07}$ & 78.64$_{\pm0.43}$ & 73.01$_{\pm0.75}$ & 65.19$_{\pm0.64}$ & 62.22$_{\pm0.57}$ & 59.58$_{\pm1.00}$ & 85.37$_{\pm0.09}$ & 83.21$_{\pm0.16}$ & 82.19$_{\pm0.17}$ & 51.77$_{\pm2.22}$ & 61.31$_{\pm0.25}$ & \textbf{\textcolor{blue}{63.34$_{\pm0.19}$}} \\
FedTGP & 79.50$_{\pm1.40}$ & 78.47$_{\pm0.64}$ & 74.58$_{\pm0.35}$ & 61.10$_{\pm0.72}$ & 59.47$_{\pm0.77}$ & 59.30$_{\pm0.48}$ & 85.01$_{\pm0.20}$ & \textbf{\textcolor{blue}{83.36$_{\pm0.13}$}} & 81.34$_{\pm0.13}$ & 42.08$_{\pm1.93}$ & 49.43$_{\pm2.00}$ & 54.24$_{\pm0.45}$ \\
FIARSE & 77.96$_{\pm0.30}$ & 78.91$_{\pm0.28}$ & 73.96$_{\pm0.40}$ & 65.17$_{\pm0.96}$ & \textbf{\textcolor{blue}{63.43$_{\pm0.46}$}} & \textbf{\textcolor{blue}{60.45$_{\pm0.68}$}} & \textbf{\textcolor{blue}{85.38$_{\pm0.22}$}} & 83.24$_{\pm0.11}$ & \textbf{\textcolor{blue}{82.32$_{\pm0.14}$}} & 60.10$_{\pm0.32}$ & 61.45$_{\pm0.26}$ & 61.49$_{\pm0.10}$ \\
FedPPN & 76.69$_{\pm0.52}$ & 78.46$_{\pm0.36}$ & 74.94$_{\pm0.27}$ & 64.87$_{\pm0.72}$ & 63.06$_{\pm0.51}$ & 60.26$_{\pm0.36}$ & 85.33$_{\pm0.19}$ & 82.44$_{\pm0.19}$ & 82.01$_{\pm0.10}$ & 60.14$_{\pm0.22}$ & 61.89$_{\pm0.12}$ & 62.75$_{\pm0.25}$ \\
TRUST & 79.89$_{\pm0.22}$ & 78.67$_{\pm0.26}$ & 74.76$_{\pm0.32}$ & 65.06$_{\pm0.57}$ & 63.16$_{\pm0.32}$ & 60.32$_{\pm0.45}$ & 85.28$_{\pm0.21}$ & 82.83$_{\pm0.17}$ & 82.12$_{\pm0.12}$ & \textbf{\textcolor{blue}{61.34$_{\pm0.32}$}} & \textbf{\textcolor{blue}{61.93$_{\pm0.25}$}} & 62.92$_{\pm0.31}$ \\
\rowcolor[HTML]{D7F6FF} \textbf{FedGKC} & \textbf{\textcolor{red}{83.42$_{\pm0.45}$}} & \textbf{\textcolor{red}{82.71$_{\pm0.62}$}} & \textbf{\textcolor{red}{76.38$_{\pm0.38}$}} & \textbf{\textcolor{red}{68.19$_{\pm0.46}$}} & \textbf{\textcolor{red}{66.15$_{\pm0.73}$}} & \textbf{\textcolor{red}{63.40$_{\pm0.67}$}} & \textbf{\textcolor{red}{87.54$_{\pm0.16}$}} & \textbf{\textcolor{red}{85.48$_{\pm0.25}$}} & \textbf{\textcolor{red}{84.42$_{\pm0.11}$}} & \textbf{\textcolor{red}{63.47$_{\pm1.90}$}} & \textbf{\textcolor{red}{62.58$_{\pm1.60}$}} & \textbf{\textcolor{red}{65.29$_{\pm4.98}$}} \\
\hline
\rowcolor{gray!20} & \multicolumn{3}{c}{CS} & \multicolumn{3}{c}{Physics} & \multicolumn{3}{c}{Photo} & \multicolumn{3}{c}{Computer} \\
\cline{2-13} \rowcolor{gray!20} \multirow{-2}{*}{Method} & 5 & 10 & 20 & 5 & 10 & 20 & 5 & 10 & 20 & 5 & 10 & 20 \\
\hline
LG-FedAvg &  88.07$_{\pm0.44}$ & 87.16$_{\pm0.31}$ & 82.69$_{\pm0.57}$ & \textbf{\textcolor{blue}{94.26$_{\pm0.28}$}} & 91.79$_{\pm0.35}$ & 92.24$_{\pm0.46}$ & 90.41$_{\pm0.53}$ & 86.81$_{\pm0.40}$ & 85.11$_{\pm0.48}$ & 87.42$_{\pm0.42}$ & 85.99$_{\pm0.37}$ & 83.79$_{\pm0.55}$ \\
FedProto & 84.09$_{\pm0.38}$ & 83.49$_{\pm0.46}$ & 81.24$_{\pm0.24}$ & 92.87$_{\pm0.11}$ & 92.54$_{\pm0.11}$ & 91.81$_{\pm0.10}$ & 71.93$_{\pm3.06}$ & 75.38$_{\pm1.52}$ & 79.37$_{\pm1.22}$ & 57.61$_{\pm2.55}$ & 75.55$_{\pm2.04}$ & 71.35$_{\pm0.74}$ \\
FML & 87.34$_{\pm0.42}$ & 85.93$_{\pm0.36}$ & 83.34$_{\pm0.49}$ & 93.03$_{\pm0.27}$ & 92.58$_{\pm0.33}$ & 92.11$_{\pm0.39}$ & 89.80$_{\pm0.44}$ & 86.84$_{\pm0.41}$ & 79.97$_{\pm0.52}$ & 87.64$_{\pm0.47}$ & 83.08$_{\pm0.46}$ & 78.18$_{\pm0.40}$ \\
FedKD & \textbf{\textcolor{blue}{88.47$_{\pm0.15}$}} & 86.46$_{\pm0.31}$ & 83.43$_{\pm0.28}$ & 93.21$_{\pm0.36}$ & 92.81$_{\pm0.20}$ & 91.86$_{\pm0.12}$ & 89.90$_{\pm4.32}$ & 84.96$_{\pm1.18}$ & 86.74$_{\pm0.58}$ & 78.53$_{\pm4.60}$ & 83.69$_{\pm0.85}$ & 78.47$_{\pm1.34}$ \\
FedTGP & 85.88$_{\pm0.15}$ & 84.99$_{\pm0.24}$ & 82.29$_{\pm0.32}$ & 93.42$_{\pm0.19}$ & 93.12$_{\pm0.12}$ & \textbf{\textcolor{blue}{92.26$_{\pm0.10}$}} & 75.31$_{\pm3.61}$ & 81.04$_{\pm1.19}$ & 80.13$_{\pm0.30}$ & 67.30$_{\pm4.97}$ & 73.76$_{\pm1.58}$ & 71.85$_{\pm1.64}$ \\
FIARSE & 87.33$_{\pm0.23}$ & \textbf{\textcolor{blue}{86.75$_{\pm0.28}$}} & \textbf{\textcolor{blue}{84.26$_{\pm0.33}$}} & 93.92$_{\pm0.11}$ & 92.97$_{\pm0.08}$ & 92.20$_{\pm0.05}$ & \textbf{\textcolor{blue}{90.84$_{\pm0.38}$}} & 89.49$_{\pm0.22}$ & 88.38$_{\pm0.15}$ & 87.71$_{\pm0.30}$ & \textbf{\textcolor{blue}{86.47$_{\pm0.19}$}} & 84.52$_{\pm0.41}$ \\
FedPPN & 86.67$_{\pm0.21}$ & 86.59$_{\pm0.20}$ & 84.10$_{\pm0.25}$ & 93.57$_{\pm0.11}$ & \textbf{\textcolor{blue}{93.12$_{\pm0.08}$}} & 91.83$_{\pm0.15}$ & 90.38$_{\pm0.35}$ & \textbf{\textcolor{blue}{90.33$_{\pm0.28}$}} & \textbf{\textcolor{blue}{88.48$_{\pm0.20}$}} & \textbf{\textcolor{blue}{87.71$_{\pm0.17}$}} & 86.09$_{\pm0.53}$ & \textbf{\textcolor{blue}{84.91$_{\pm0.24}$}} \\
TRUST & 87.53$_{\pm0.32}$ & 85.78$_{\pm0.24}$ & 83.32$_{\pm0.31}$ & 94.43$_{\pm0.22}$ & 92.56$_{\pm0.15}$ & 92.05$_{\pm0.21}$ & 89.45$_{\pm0.24}$ & 82.44$_{\pm0.19}$ & 89.45$_{\pm0.18}$ & 84.32$_{\pm0.15}$ & 83.57$_{\pm0.25}$ & 82.94$_{\pm0.32}$ \\
\rowcolor[HTML]{D7F6FF} \textbf{FedGKC} & \textbf{\textcolor{red}{89.62$_{\pm0.15}$}} & \textbf{\textcolor{red}{88.31$_{\pm0.27}$}} & \textbf{\textcolor{red}{86.04$_{\pm0.14}$}} & \textbf{\textcolor{red}{95.71$_{\pm0.27}$}} & \textbf{\textcolor{red}{94.75$_{\pm0.15}$}} & \textbf{\textcolor{red}{93.56$_{\pm0.26}$}} & \textbf{\textcolor{red}{93.19$_{\pm0.67}$}} & \textbf{\textcolor{red}{91.65$_{\pm0.32}$}} & \textbf{\textcolor{red}{89.76$_{\pm0.43}$}} & \textbf{\textcolor{red}{88.57$_{\pm0.34}$}} & \textbf{\textcolor{red}{87.90$_{\pm0.32}$}} & \textbf{\textcolor{red}{86.84$_{\pm0.32}$}} \\
\Xhline{1pt}
\end{tabular}}
\vspace{-2mm}
\end{table*}

\begin{table*}[t]
\caption{Performance comparison in \textcolor{magenta}{Varying Model Scales}. The best and sub-best results are marked in \textbf{\textcolor{red}{Red}} and \textbf{\textcolor{blue}{Blue}}.}
\vspace{-3mm}
\label{tab:performance2}
\resizebox{\linewidth}{!}{
\begin{tabular}{lcccccccccccc}
\Xhline{1pt}
\rowcolor{gray!20} & \multicolumn{3}{c}{Cora} & \multicolumn{3}{c}{CiteSeer} & \multicolumn{3}{c}{PubMed} & \multicolumn{3}{c}{ogbn-arxiv} \\
\cline{2-13} \rowcolor{gray!20} \multirow{-2}{*}{Method} & 5 & 10 & 20 & 5 & 10 & 20 & 5 & 10 & 20 & 5 & 10 & 20 \\
\hline
LG-FedAvg & 79.50$_{\pm0.42}$ & 79.90$_{\pm0.37}$ & 73.43$_{\pm0.55}$ & 66.16$_{\pm0.44}$ & \textbf{\textcolor{blue}{65.47$_{\pm0.39}$}} & 58.62$_{\pm0.52}$ & 84.48$_{\pm0.41}$ & 83.17$_{\pm0.46}$ & 80.70$_{\pm0.38}$ & 62.24$_{\pm0.47}$ & 65.05$_{\pm0.51}$ & 65.32$_{\pm0.43}$ \\
FedProto & 77.59$_{\pm0.61}$ & 78.19$_{\pm0.57}$ & 72.55$_{\pm0.21}$ & 62.08$_{\pm0.91}$ & 57.70$_{\pm1.47}$ & 57.03$_{\pm0.33}$ & 84.71$_{\pm3.32}$ & 83.08$_{\pm0.30}$ & 81.27$_{\pm0.27}$ & 50.06$_{\pm1.34}$ & 53.92$_{\pm0.91}$ & 57.04$_{\pm0.25}$ \\
FML & \textbf{\textcolor{blue}{80.42$_{\pm0.48}$}} & 79.18$_{\pm0.39}$ & 74.23$_{\pm0.44}$ & 66.23$_{\pm0.36}$ & 64.30$_{\pm0.41}$ & \textbf{\textcolor{blue}{61.10$_{\pm0.52}$}} & 84.76$_{\pm0.47}$ & 83.13$_{\pm0.42}$ & 81.43$_{\pm0.38}$ & 54.89$_{\pm0.56}$ & 62.43$_{\pm0.49}$ & 64.62$_{\pm0.45}$ \\
FedKD & 80.23$_{\pm0.47}$ & \textbf{\textcolor{blue}{79.81$_{\pm0.56}$}} & 74.13$_{\pm0.39}$ & 66.46$_{\pm0.94}$ & 65.92$_{\pm1.26}$ & 60.30$_{\pm0.58}$ & \textbf{\textcolor{blue}{85.08$_{\pm2.51}$}} & 83.25$_{\pm0.16}$ & \textbf{\textcolor{blue}{82.21$_{\pm0.66}$}} & 59.71$_{\pm0.87}$ & 61.06$_{\pm6.22}$ & 63.68$_{\pm5.00}$ \\
FedTGP & 78.23$_{\pm0.75}$ & 78.65$_{\pm0.66}$ & 73.25$_{\pm0.44}$ & 63.12$_{\pm0.61}$ & 58.60$_{\pm0.73}$ & 58.55$_{\pm0.65}$ & 84.43$_{\pm2.95}$ & \textbf{\textcolor{blue}{83.32$_{\pm0.16}$}} & 81.12$_{\pm0.33}$ & 53.02$_{\pm2.57}$ & 58.10$_{\pm0.72}$ & 59.23$_{\pm0.18}$ \\
FIARSE & 78.78$_{\pm0.24}$ & 79.18$_{\pm0.44}$ & \textbf{\textcolor{blue}{74.90$_{\pm0.31}$}} & 66.17$_{\pm0.52}$ & 61.36$_{\pm0.96}$ & 59.90$_{\pm0.64}$ & 79.61$_{\pm3.11}$ & 78.35$_{\pm0.16}$ & 80.38$_{\pm0.35}$ & 60.76$_{\pm0.54}$ & 61.77$_{\pm0.40}$ & 62.03$_{\pm0.16}$ \\
FedPPN & 78.42$_{\pm0.14}$ & 78.46$_{\pm0.50}$ & 74.62$_{\pm0.28}$ & \textbf{\textcolor{blue}{66.55$_{\pm0.64}$}} & 60.81$_{\pm0.93}$ & 59.66$_{\pm0.62}$ & 83.43$_{\pm0.51}$ & 78.48$_{\pm0.27}$ & 80.11$_{\pm0.49}$ & 60.50$_{\pm0.20}$ & 61.86$_{\pm0.24}$ & 61.77$_{\pm0.27}$ \\
TRUST & 78.73$_{\pm0.32}$ & 78.14$_{\pm0.18}$ & 74.04$_{\pm0.31}$ & 65.15$_{\pm0.52}$ & 63.51$_{\pm0.31}$ & 60.43$_{\pm0.27}$ & 84.94$_{\pm0.27}$ & 83.25$_{\pm0.28}$ & 82.14$_{\pm0.13}$ & \textbf{\textcolor{blue}{62.34$_{\pm0.24}$}} & \textbf{\textcolor{blue}{65.12$_{\pm0.27}$}} & \textbf{\textcolor{blue}{65.34$_{\pm0.45}$}} \\
\rowcolor[HTML]{D7F6FF} \textbf{FedGKC} & \textbf{\textcolor{red}{82.35$_{\pm0.32}$}} & \textbf{\textcolor{red}{81.25$_{\pm0.51}$}} & \textbf{\textcolor{red}{77.29$_{\pm0.34}$}} & \textbf{\textcolor{red}{68.79$_{\pm0.37}$}} & \textbf{\textcolor{red}{67.27$_{\pm0.74}$}} & \textbf{\textcolor{red}{63.45$_{\pm0.43}$}} & \textbf{\textcolor{red}{87.35$_{\pm0.23}$}} & \textbf{\textcolor{red}{86.94$_{\pm0.36}$}} & \textbf{\textcolor{red}{84.61$_{\pm0.31}$}} & \textbf{\textcolor{red}{64.25$_{\pm0.32}$}} & \textbf{\textcolor{red}{67.65$_{\pm0.88}$}} & \textbf{\textcolor{red}{68.09$_{\pm1.04}$}} \\
\hline
\rowcolor{gray!20} & \multicolumn{3}{c}{CS} & \multicolumn{3}{c}{Physics} & \multicolumn{3}{c}{Photo} & \multicolumn{3}{c}{Computer} \\
\cline{2-13} \rowcolor{gray!20} \multirow{-2}{*}{Method} & 5 & 10 & 20 & 5 & 10 & 20 & 5 & 10 & 20 & 5 & 10 & 20 \\
\hline
LG-FedAvg & \textbf{\textcolor{blue}{90.64$_{\pm0.47}$}} & \textbf{\textcolor{blue}{88.04$_{\pm0.42}$}} & 85.32$_{\pm0.54}$ & 94.42$_{\pm0.33}$ & 93.87$_{\pm0.39}$ & \textbf{\textcolor{blue}{93.45$_{\pm0.28}$}} & 89.96$_{\pm0.52}$ & 87.91$_{\pm0.44}$ & 85.59$_{\pm0.49}$ & 85.77$_{\pm0.41}$ & \textbf{\textcolor{blue}{85.53$_{\pm0.46}$}} & \textbf{\textcolor{blue}{84.76$_{\pm0.50}$}} \\
FedProto & 83.95$_{\pm0.28}$ & 83.08$_{\pm0.08}$ & 81.80$_{\pm0.38}$ & 93.12$_{\pm0.15}$ & 92.42$_{\pm0.13}$ & 91.86$_{\pm0.08}$ & 71.35$_{\pm3.39}$ & 73.48$_{\pm2.00}$ & 80.17$_{\pm0.57}$ & 73.38$_{\pm0.48}$ & 70.88$_{\pm1.73}$ & 73.44$_{\pm0.79}$ \\
FML & 89.23$_{\pm0.46}$ & 86.65$_{\pm0.42}$ & 85.76$_{\pm0.51}$ & \textbf{\textcolor{blue}{94.60$_{\pm0.29}$}} & 93.91$_{\pm0.36}$ & 92.11$_{\pm0.33}$ & 82.65$_{\pm0.44}$ & 87.68$_{\pm0.47}$ & 85.50$_{\pm0.39}$ & 86.19$_{\pm0.48}$ & 85.28$_{\pm0.40}$ & 84.60$_{\pm0.45}$ \\
FedKD & 89.48$_{\pm0.14}$ & 87.03$_{\pm0.14}$ & 85.88$_{\pm0.11}$ & 94.51$_{\pm0.33}$ & \textbf{\textcolor{blue}{94.00$_{\pm0.23}$}} & 92.76$_{\pm0.08}$ & 84.66$_{\pm0.42}$ & 85.70$_{\pm0.63}$ & 86.01$_{\pm0.61}$ & 77.84$_{\pm3.61}$ & 80.35$_{\pm2.23}$ & 81.90$_{\pm1.08}$ \\
FedTGP & 86.84$_{\pm0.41}$ & 85.49$_{\pm0.28}$ & 83.81$_{\pm0.11}$ & 93.92$_{\pm0.15}$ & 92.57$_{\pm0.05}$ & 93.03$_{\pm0.07}$ & 71.42$_{\pm3.05}$ & 73.72$_{\pm1.67}$ & 81.81$_{\pm0.61}$ & 66.73$_{\pm3.67}$ & 67.63$_{\pm0.53}$ & 70.27$_{\pm1.28}$ \\
FIARSE & 87.77$_{\pm0.33}$ & 85.80$_{\pm0.13}$ & \textbf{\textcolor{blue}{85.94$_{\pm0.17}$}} & 93.40$_{\pm0.09}$ & 93.69$_{\pm0.06}$ & 92.28$_{\pm0.14}$ & 90.19$_{\pm0.35}$ & \textbf{\textcolor{blue}{89.34$_{\pm0.16}$}} & \textbf{\textcolor{blue}{86.50$_{\pm0.50}$}} & \textbf{\textcolor{blue}{86.85$_{\pm0.42}$}} & 85.02$_{\pm0.26}$ & 83.73$_{\pm0.35}$ \\
FedPPN & 87.74$_{\pm0.29}$ & 85.29$_{\pm0.26}$ & 85.31$_{\pm0.30}$ & 93.38$_{\pm0.21}$ & 93.57$_{\pm0.08}$ & 92.14$_{\pm0.08}$ & \textbf{\textcolor{blue}{90.52$_{\pm0.29}$}} & 89.30$_{\pm0.15}$ & 86.47$_{\pm0.47}$ & 86.73$_{\pm0.47}$ & 84.51$_{\pm1.11}$ & 83.53$_{\pm0.16}$ \\
TRUST & 88.94$_{\pm0.26}$ & 87.35$_{\pm0.24}$ & 84.54$_{\pm0.26}$ & 94.23$_{\pm0.51}$ & 93.35$_{\pm0.37}$ & 92.21$_{\pm0.25}$ & 88.51$_{\pm0.34}$ & 87.81$_{\pm0.43}$ & 85.85$_{\pm0.16}$ & 85.84$_{\pm0.28}$ & 84.63$_{\pm0.26}$ & 83.76$_{\pm0.42}$ \\
\rowcolor[HTML]{D7F6FF} \textbf{FedGKC} & \textbf{\textcolor{red}{92.58$_{\pm0.42}$}} & \textbf{\textcolor{red}{90.91$_{\pm0.18}$}} & \textbf{\textcolor{red}{88.50$_{\pm0.15}$}} & \textbf{\textcolor{red}{95.35$_{\pm0.23}$}} & \textbf{\textcolor{red}{94.78$_{\pm0.15}$}} & \textbf{\textcolor{red}{93.84$_{\pm0.15}$}} & \textbf{\textcolor{red}{92.87$_{\pm0.28}$}} & \textbf{\textcolor{red}{90.54$_{\pm0.31}$}} & \textbf{\textcolor{red}{89.86$_{\pm0.36}$}} & \textbf{\textcolor{red}{89.54$_{\pm0.31}$}} & \textbf{\textcolor{red}{87.64$_{\pm0.40}$}} & \textbf{\textcolor{red}{86.48$_{\pm0.32}$}} \\
\Xhline{1pt} 
\end{tabular}}
\vspace{-2mm}
\end{table*}

\begin{table}[t]
\caption{Comparison of our method with other federated methods with 10 clients in non-heterogeneous settings.}
\vspace{-3mm}
\label{tab:nonht}
\resizebox{\linewidth}{!}{
\begin{tabular}{lccccc}
\Xhline{1pt} 
\rowcolor{gray!20} & CS & Physics & Photo & Computer & Avg. \\
\hline
FedAvg & 81.06$_{\pm0.62}$ & 92.71$_{\pm0.47}$ & 78.02$_{\pm0.53}$ & 66.90$_{\pm0.71}$ & 79.67$_{\pm0.58}$ \\
MOON & 85.02$_{\pm0.44}$ & 94.37$_{\pm0.68}$ & 80.37$_{\pm0.59}$ & 69.50$_{\pm0.41}$ & 82.32$_{\pm0.52}$ \\
FedDC & 83.71$_{\pm0.71}$ & 93.04$_{\pm0.55}$ & 72.99$_{\pm0.66}$ & 63.35$_{\pm0.79}$ & 78.27$_{\pm0.63}$ \\
FedSage+ & 84.83$_{\pm0.39}$ & 92.96$_{\pm0.48}$ & 79.21$_{\pm0.42}$ & 68.84$_{\pm0.57}$ & 81.46$_{\pm0.46}$ \\
Fed-Pub & 88.58$_{\pm0.32}$ & 92.11$_{\pm0.27}$ & 89.17$_{\pm0.36}$ & 86.54$_{\pm0.29}$ & 89.10$_{\pm0.31}$ \\
FGSSL & 86.47$_{\pm0.41}$ & 92.66$_{\pm0.33}$ & 87.94$_{\pm0.29}$ & 85.50$_{\pm0.37}$ & 88.14$_{\pm0.35}$ \\
FedGTA & 87.37$_{\pm0.28}$ & 93.80$_{\pm0.44}$ & 87.74$_{\pm0.31}$ & 84.58$_{\pm0.39}$ & 88.37$_{\pm0.35}$ \\
AdaFGL & 85.77$_{\pm0.55}$ & 93.88$_{\pm0.47}$ & 86.30$_{\pm0.52}$ & 83.29$_{\pm0.61}$ & 87.31$_{\pm0.54}$ \\
\rowcolor[HTML]{D7F6FF} \textbf{FedGKC} & \textbf{88.73$_{\pm0.33}$} & \textbf{94.47$_{\pm0.29}$} & \textbf{90.30$_{\pm0.27}$} & \textbf{87.16$_{\pm0.31}$} & \textbf{90.17$_{\pm0.30}$} \\
\Xhline{1pt}
\end{tabular}}
\vspace{-2mm}
\end{table}

\subsection{Performance Comparison}
    To answer \textbf{Q1}, we conduct comprehensive experiments under heterogeneous and non-heterogeneous settings.
    Overall, FedGKC exhibits consistent improvements over all selected MHtFL baselines.
    \textbf{(1) Clients with Different Model Architectures.}
    Table~\ref{tab:performance} summarizes results when clients adopt different model architectures
    FedGKC achieves the best or second-best performance across all datasets and client configs, surpassing KD and prototype-based methods.
    Two strong partial-model strategies (i.e., FIARSE, FedPPN) exhibit competitive results due to their effective submodel selection and parameter coverage.
    However, their performances are less stable when architectural heterogeneity increases, as partial extraction/mapping struggles to preserve topology-aware signals across diverse GNN designs.
    In contrast, FedGKC decouples cross-client exchange via copilot models and integrates topology-aware distillation, resulting in superior overall accuracy and more robust transfer.
    \textbf{(2) Clients with Different Model Scales.}
    In the second scenario, we consider the variations in model scales due to differences in the computational capabilities of various client devices.
    As shown in Table~\ref{tab:performance2}, FedGKC demonstrates superior performance on the test set compared to other methods, with an average of 4.40\% higher accuracy than FML and 4.22\% higher accuracy than FedKD.
    It confirms that FedGKC can not only adapt to the heterogeneity of simple model architectures, but can also be extended to models of different scales, and performs well.
    \textbf{(3) Comparative Experiments under Non-Heterogeneous Settings.}
    In addition to the experiments under heterogeneous settings, we also conduct comparative experiments in non-heterogeneous settings to evaluate the performance of FedGKC in traditional federated learning environments.
    The results in Table~\ref{tab:nonht} show that despite being specifically designed for heterogeneous client scenarios, FedGKC performs exceptionally well even in non-heterogeneous settings, outperforming the baseline approach FedAvg by 10.5\% on average.
    FedGKC even surpasses federated learning methods tailored for homogeneous clients by an average of 1.07\% over the suboptimal method Fed-Pub, which further validates the generality and robustness of FedGKC.

\begin{table*}[t]
\caption{The experimental results by incorporating the proposed key components into the baseline. "SMKD" indicates Self-Mutual Knowledge Distillation, and "KAMA" indicates Knowledge-Aware Model Aggregation.}
\vspace{-3mm}
\label{tab:compoent}
\resizebox{\linewidth}{!}{
\begin{tabular}{cccccccccccccc}
\Xhline{1pt}
\rowcolor{gray!20} & & \multicolumn{4}{c}{Cora} & \multicolumn{4}{c}{CiteSeer} & \multicolumn{4}{c}{PubMed} \\
\hline
SMKD & KAMA & Client 5 & Client 10 & Client 20 & Avg. & Client 5 & Client 10 & Client 20 & Avg. & Client 5 & Client 10 & Client 20 & Avg. \\
\hline
& & 80.50$_{\pm0.41}$ & 79.54$_{\pm0.36}$ & 75.21$_{\pm0.58}$ & 78.42$_{\pm0.45}$ & 64.74$_{\pm0.47}$ & 62.73$_{\pm0.39}$ & 60.90$_{\pm0.55}$ & 62.82$_{\pm0.47}$ & 84.64$_{\pm0.33}$ & 83.15$_{\pm0.28}$ & 81.87$_{\pm0.42}$ & 83.22$_{\pm0.34}$ \\
\checkmark & & 81.41$_{\pm0.45}$ & 80.44$_{\pm0.37}$ & 74.50$_{\pm0.52}$ & 78.79$_{\pm0.45}$ & 64.44$_{\pm0.46}$ & 64.29$_{\pm0.44}$ & 60.88$_{\pm0.49}$ & 63.20$_{\pm0.46}$ & 84.54$_{\pm0.41}$ & 83.37$_{\pm0.35}$ & \textbf{82.51$_{\pm0.40}$} & 83.47$_{\pm0.39}$ \\
& \checkmark & 81.24$_{\pm0.39}$ & 79.73$_{\pm0.41}$ & 75.03$_{\pm0.46}$ & 78.67$_{\pm0.42}$ & 65.04$_{\pm0.42}$ & 63.11$_{\pm0.38}$ & 61.03$_{\pm0.51}$ & 63.06$_{\pm0.44}$ & 85.05$_{\pm0.36}$ & 83.20$_{\pm0.33}$ & 82.27$_{\pm0.40}$ & 83.51$_{\pm0.36}$ \\
\rowcolor[HTML]{D7F6FF} \checkmark & \checkmark & \textbf{81.42$_{\pm0.45}$} & \textbf{80.71$_{\pm0.62}$} & \textbf{75.38$_{\pm0.38}$} & \textbf{79.17$_{\pm0.48}$} & \textbf{67.19$_{\pm0.46}$} & \textbf{65.15$_{\pm0.73}$} & \textbf{61.40$_{\pm0.67}$} & \textbf{64.58$_{\pm0.62}$} & \textbf{85.54$_{\pm0.16}$} & \textbf{83.48$_{\pm0.25}$} & 82.42$_{\pm0.11}$ & \textbf{83.81$_{\pm0.17}$} \\
\Xhline{1pt}
\end{tabular}}
\vspace{-3mm}
\end{table*}

\subsection{Ablation Study}
\label{sec: Ablation Study}
    To address \textbf{Q2}, we perform ablation experiments on the two key modules of FedGKC: Self-Mutual Knowledge Distillation (SMKD) and Knowledge-Aware Model Aggregation (KAMA).
    Table~\ref{tab:compoent} reports the results when each module is applied independently.
    Specifically, when SMKD is used alone, the server employs a sample-size–based aggregation strategy; when KAMA is used alone, clients adhere to a conventional dual-model architecture.
    The results show that SMKD provides notable improvements on Cora and CiteSeer, while KAMA contributes more strongly on PubMed, suggesting that KAMA is particularly advantageous for larger-scale datasets.
    Combining both modules yields the best overall performance across all benchmarks.
    
\begin{figure}[t]
\centering
\includegraphics[width=\linewidth]{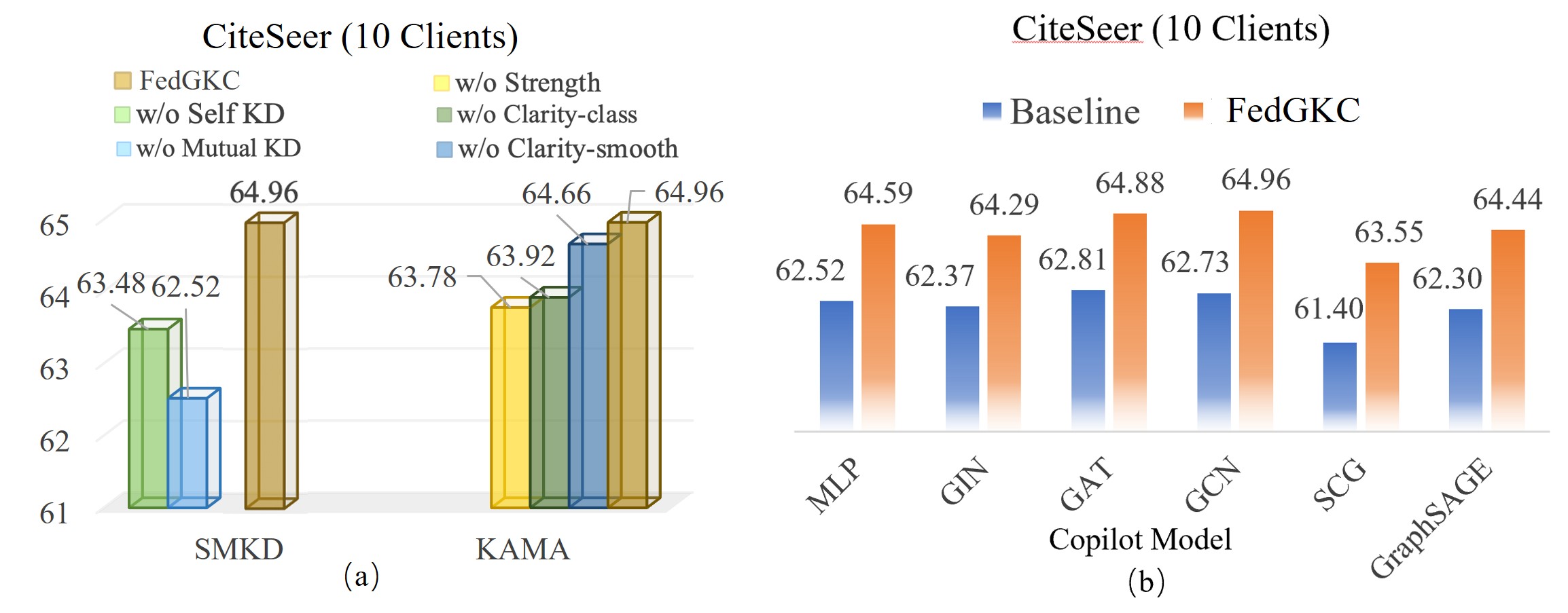}
\vspace{-6mm}
\caption{(a) Ablation experiments of SMKD and KAMA. (b) Coarse local–copilot Alignment vs. full FedGKC using different model Architectures on the copilot model.}
\vspace{-5mm}
\label{fig_component}
\end{figure}

    To further disentangle the contributions of individual mechanisms, we conduct a fine-grained ablation on the CiteSeer with 10 clients, as shown in Fig.~\ref{fig_component}(a).
    For the SMKD module, we separately remove the self-distillation and mutual distillation components.
    Results indicate that eliminating self-distillation weakens the model’s ability to optimize its own representations, while removing mutual distillation forces clients to rely solely on local data, thereby hindering global knowledge assimilation and degrading overall performance.
    For the KAMA, we ablate the three factors forming knowledge weights: knowledge strength, class-level knowledge clarity (the first term in Eq.~(\ref{equ:clarity})), and association clarity (the second term in Eq.~(\ref{equ:clarity})).
    We find that knowledge strength plays the most decisive role in aggregation, whereas the others exert relatively smaller but positive effects, enhancing both stability and accuracy.

\subsection{Generalization and Sensitivity Analysis}
\label{sec: Generalization and Sensitivity Analysis}
    To answer \textbf{Q3}, we investigate the effectiveness of mutual distillation module when deploying varying architectures on copilot model in Fig.~\ref{fig_component}(b).
    Baseline refers to removal of carefully designed loss objectives and KAMA-related mechanism while applying various model architectures on copilot model.
    As results demonstrate, full FedGKC demonstrates significant performance improvements across choices of most widely adopted model architectures, further validating the flexibility and generalizability of FedGKC's design.

\begin{figure}[t]
\centering
\includegraphics[width=\linewidth]{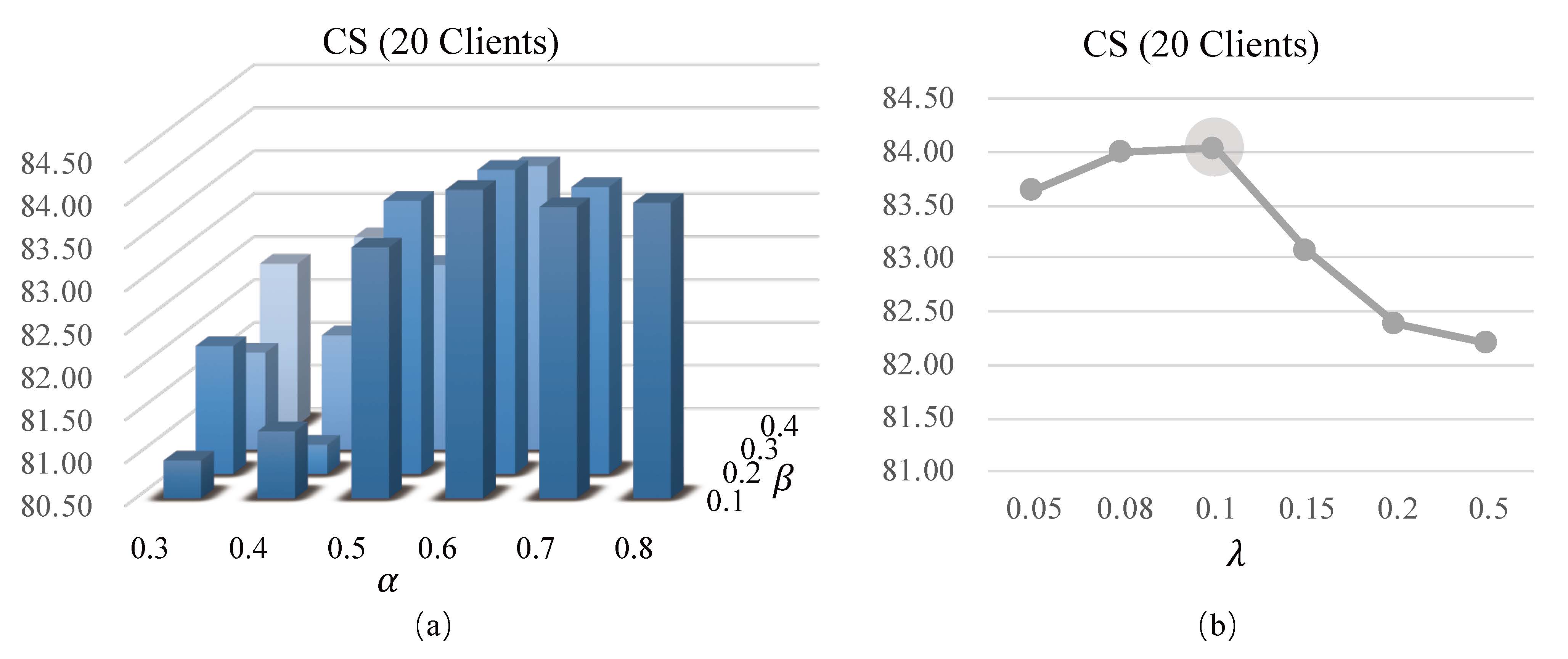}
\vspace{-7.5mm}
\caption{Hyperparameters analysis: (a) Different combinations between parameters $\alpha$ and $\beta$. (b) Different $\lambda$ values.}
\vspace{-4.5mm}
\label{fig_hypara}
\end{figure}

    To answer \textbf{Q4}, we evaluate the performance of FedGKC using 20 clients from the CS dataset across various hyperparameter settings.
    Specifically, we explore various combinations of the trade-off parameters $\alpha$ and $\beta$ in Eq.~(\ref{equ:copmodel}) and Eq.~(\ref{equ:localmutu}).
    The results in Fig.~\ref{fig_hypara}(a) indicate that FedGKC performs optimally with $\alpha$ set between 0.6 and 0.7, and $\beta$ maintained within the range of 0.1 to 0.2.
    Accordingly, we finalize our hyperparameters with $\alpha=0.6$ and $\beta=0.2$. Additionally, we conduct a sensitivity analysis on $\lambda$ in Eq.~(\ref{equ:clarity}), which represents the influence of over-smoothing on the clarity of node knowledge.
    From Fig.~\ref{fig_hypara}(b), we find that setting $\lambda$ to 0.1 effectively enhances model performance, whereas excessively high values lead to performance deterioration. It underscores the importance of balancing node similarity and the over-smoothing issue.

\section{Conclusion}
\label{sec: Conclusion}
    This paper introduces FedGKC, a knowledge-driven FGL framework resolving the model-heterogeneity challenge. By coupling mutual knowledge distillation between each local model and globally trained copilot model and performing knowledge-aware aggregation on the server, FedGKC enables clients with different model architectures to collaborate without sharing raw data. Empirically and theoretically, FedGKC demonstrates robust performance and stable optimization while keeping communication efficient.
    Future works include improving knowledge transfer on graphs, and refining mechanisms to further balance efficiency and accuracy.

\clearpage

\bibliographystyle{ACM-Reference-Format}
\bibliography{sample-base}

@String{Computing = "Computing" }

@String{Computer = "{IEEE} Computer" }

@String{Springer = "Springer-Verlag" }

@inproceedings{feng2022freekd,
  title={Freekd: Free-direction knowledge distillation for graph neural networks},
  author={Feng, Kaituo and Li, Changsheng and Yuan, Ye and Wang, Guoren},
  booktitle={Proceedings of the ACM SIGKDD Conference on Knowledge Discovery and Data Mining, KDD},
  year={2022}
}

@inproceedings{zhang2020rdd,
  title={Reliable data distillation on graph convolutional network},
  author={Zhang, Wentao and Miao, Xupeng and Shao, Yingxia and Jiang, Jiawei and Chen, Lei and Ruas, Olivier and Cui, Bin},
  booktitle={Proceedings of the ACM on Management of Data, SIGMOD},
  year={2020}
}

@article{zhang2021fedsage,
  title={Subgraph federated learning with missing neighbor generation},
  author={Zhang, Ke and Yang, Carl and Li, Xiaoxiao and Sun, Lichao and Yiu, Siu Ming},
  journal={Advances in Neural Information Processing Systems, NeurIPS},
  year={2021}
}

@inproceedings{tan2023fedstar,
  title={Federated learning on non-iid graphs via structural knowledge sharing},
  author={Tan, Yue and Liu, Yixin and Long, Guodong and Jiang, Jing and Lu, Qinghua and Zhang, Chengqi},
  booktitle={Proceedings of the AAAI Conference on Artificial Intelligence, AAAI},
  year={2023}
}

@inproceedings{zhu2021fedgen,
  title={Data-free knowledge distillation for heterogeneous federated learning},
  author={Zhu, Zhuangdi and Hong, Junyuan and Zhou, Jiayu},
  booktitle={International Conference on Machine Learning, ICML},
  year={2021},
  organization={PMLR}
}

@inproceedings{wu2019sgc,
  title={Simplifying graph convolutional networks},
  author={Wu, Felix and Souza, Amauri and Zhang, Tianyi and Fifty, Christopher and Yu, Tao and Weinberger, Kilian},
  booktitle={International Conference on Machine Learning, ICML},
  year={2019}
}

@inproceedings{CoraDataset,
  title={Revisiting semi-supervised learning with graph embeddings},
  author={Yang, Zhilin and Cohen, William and Salakhudinov, Ruslan},
  booktitle={International conference on machine learning},
  pages={40--48},
  year={2016},
  organization={PMLR}
}

@article{CSandPhyDataset,
  title={Pitfalls of graph neural network evaluation},
  author={Shchur, Oleksandr and Mumme, Maximilian and Bojchevski, Aleksandar and G{\"u}nnemann, Stephan},
  journal={arXiv preprint arXiv:1811.05868},
  year={2018}
}

@inproceedings{mcauley2015ComputDataset,
  title={Image-based recommendations on styles and substitutes},
  author={McAuley, Julian and Targett, Christopher and Shi, Qinfeng and Van Den Hengel, Anton},
  booktitle={Proceedings of the 38th international ACM SIGIR conference on research and development in information retrieval},
  pages={43--52},
  year={2015}
}

@article{ogbnarxiv,
  title={Open graph benchmark: Datasets for machine learning on graphs},
  author={Hu, Weihua and Fey, Matthias and Zitnik, Marinka and Dong, Yuxiao and Ren, Hongyu and Liu, Bowen and Catasta, Michele and Leskovec, Jure},
  journal={Advances in neural information processing systems},
  volume={33},
  pages={22118--22133},
  year={2020}
}

@article{liang2020LG,
  title={Think locally, act globally: Federated learning with local and global representations},
  author={Liang, Paul Pu and Liu, Terrance and Ziyin, Liu and Allen, Nicholas B and Auerbach, Randy P and Brent, David and Salakhutdinov, Ruslan and Morency, Louis-Philippe},
  journal={arXiv preprint arXiv:2001.01523},
  year={2020}
}

@inproceedings{yi2023fedgh,
  title={FedGH: Heterogeneous federated learning with generalized global header},
  author={Yi, Liping and Wang, Gang and Liu, Xiaoguang and Shi, Zhuan and Yu, Han},
  booktitle={Proceedings of the 31st ACM International Conference on Multimedia},
  pages={8686--8696},
  year={2023}
}

@inproceedings{tan2022fedproto,
  title={Fedproto: Federated prototype learning across heterogeneous clients},
  author={Tan, Yue and Long, Guodong and Liu, Lu and Zhou, Tianyi and Lu, Qinghua and Jiang, Jing and Zhang, Chengqi},
  booktitle={Proceedings of the AAAI Conference on Artificial Intelligence},
  volume={36},
  number={8},
  pages={8432--8440},
  year={2022}
}

@inproceedings{zhang2024fedtgp,
  title={Fedtgp: Trainable global prototypes with adaptive-margin-enhanced contrastive learning for data and model heterogeneity in federated learning},
  author={Zhang, Jianqing and Liu, Yang and Hua, Yang and Cao, Jian},
  booktitle={Proceedings of the AAAI Conference on Artificial Intelligence},
  volume={38},
  number={15},
  pages={16768--16776},
  year={2024}
}

@article{shen2023fedfml,
  title={Federated mutual learning: a collaborative machine learning method for heterogeneous data, models, and objectives},
  author={Shen, Tao and Zhang, Jie and Jia, Xinkang and Zhang, Fengda and Lv, Zheqi and Kuang, Kun and Wu, Chao and Wu, Fei},
  journal={Frontiers of Information Technology \& Electronic Engineering},
  volume={24},
  number={10},
  pages={1390--1402},
  year={2023},
  publisher={Springer}
}

@article{wu2022communication,
  title={Communication-efficient federated learning via knowledge distillation},
  author={Wu, Chuhan and Wu, Fangzhao and Lyu, Lingjuan and Huang, Yongfeng and Xie, Xing},
  journal={Nature communications},
  volume={13},
  number={1},
  pages={2032},
  year={2022},
  publisher={Nature Publishing Group UK London}
}

@inproceedings{kipf2017GCN,
  title={Semi-Supervised Classification with Graph Convolutional Networks},
  author={Kipf, Thomas N and Welling, Max},
  booktitle={International Conference on Learning Representations},
  year={2017}
}

@article{hamilton2017graphsage,
  title={Inductive representation learning on large graphs},
  author={Hamilton, Will and Ying, Zhitao and Leskovec, Jure},
  journal={Advances in neural information processing systems},
  volume={30},
  year={2017}
}

@inproceedings{velivckovic2018gat,
  title={Graph Attention Networks},
  author={Veli{\v{c}}kovi{\'c}, Petar and Cucurull, Guillem and Casanova, Arantxa and Romero, Adriana and Li{\`o}, Pietro and Bengio, Yoshua},
  booktitle={International Conference on Learning Representations},
  year={2018}
}

@inproceedings{xu2018gin,
  title={How Powerful are Graph Neural Networks?},
  author={Xu, Keyulu and Hu, Weihua and Leskovec, Jure and Jegelka, Stefanie},
  booktitle={International Conference on Learning Representations},
  year={2018}
}

@article{wu2020GNN,
  title={A comprehensive survey on graph neural networks},
  author={Wu, Zonghan and Pan, Shirui and Chen, Fengwen and Long, Guodong and Zhang, Chengqi and Philip, S Yu},
  journal={IEEE transactions on neural networks and learning systems},
  volume={32},
  number={1},
  pages={4--24},
  year={2020},
  publisher={IEEE}
}

@article{kumar2022socialgraph,
  title={Influence maximization in social networks using graph embedding and graph neural network},
  author={Kumar, Sanjay and Mallik, Abhishek and Khetarpal, Anavi and Panda, Bhawani Sankar},
  journal={Information Sciences},
  volume={607},
  pages={1617--1636},
  year={2022},
  publisher={Elsevier}
}

@article{zhang2021biograph,
  title={Graph neural networks and their current applications in bioinformatics},
  author={Zhang, Xiao-Meng and Liang, Li and Liu, Lin and Tang, Ming-Jing},
  journal={Frontiers in genetics},
  volume={12},
  pages={690049},
  year={2021},
  publisher={Frontiers Media SA}
}

@inproceedings{he2020recomgraph,
  title={Lightgcn: Simplifying and powering graph convolution network for recommendation},
  author={He, Xiangnan and Deng, Kuan and Wang, Xiang and Li, Yan and Zhang, Yongdong and Wang, Meng},
  booktitle={Proceedings of the 43rd International ACM SIGIR conference on research and development in Information Retrieval},
  pages={639--648},
  year={2020}
}

@article{liu2024FGL,
  title={Federated graph neural networks: Overview, techniques, and challenges},
  author={Liu, Rui and Xing, Pengwei and Deng, Zichao and Li, Anran and Guan, Cuntai and Yu, Han},
  journal={IEEE Transactions on Neural Networks and Learning Systems},
  year={2024},
  publisher={IEEE}
}

@incollection{long2020federatedProblem,
  title={Federated learning for open banking},
  author={Long, Guodong and Tan, Yue and Jiang, Jing and Zhang, Chengqi},
  booktitle={Federated learning: privacy and incentive},
  pages={240--254},
  year={2020},
  publisher={Springer}
}

@article{xie2021GCFL,
  title={Federated graph classification over non-iid graphs},
  author={Xie, Han and Ma, Jing and Xiong, Li and Yang, Carl},
  journal={Advances in neural information processing systems},
  volume={34},
  pages={18839--18852},
  year={2021}
}

@inproceedings{ijcai2023FGSSL,
  title     = {Federated Graph Semantic and Structural Learning},
  author    = {Huang, Wenke and Wan, Guancheng and Ye, Mang and Du, Bo},
  booktitle = {Proceedings of the Thirty-Second International Joint Conference on
               Artificial Intelligence, {IJCAI-23}},
  publisher = {International Joint Conferences on Artificial Intelligence Organization},
  editor    = {Edith Elkind},
  pages     = {3830--3838},
  year      = {2023},
  month     = {8},
  note      = {Main Track},
  doi       = {10.24963/ijcai.2023/426},
  url       = {https://doi.org/10.24963/ijcai.2023/426},
}

@article{li2023fedgta,
  title={FedGTA: Topology-Aware Averaging for Federated Graph Learning},
  author={Li, Xunkai and Wu, Zhengyu and Zhang, Wentao and Zhu, Yinlin and Li, Rong-Hua and Wang, Guoren},
  journal={Proceedings of the VLDB Endowment},
  volume={17},
  number={1},
  pages={41--50},
  year={2023},
  publisher={VLDB Endowment}
}

@inproceedings{baek2023fedpub,
  title={Personalized subgraph federated learning},
  author={Baek, Jinheon and Jeong, Wonyong and Jin, Jiongdao and Yoon, Jaehong and Hwang, Sung Ju},
  booktitle={International conference on machine learning},
  pages={1396--1415},
  year={2023},
  organization={PMLR}
}

@article{zhang2021parameterized,
  title={Parameterized knowledge transfer for personalized federated learning},
  author={Zhang, Jie and Guo, Song and Ma, Xiaosong and Wang, Haozhao and Xu, Wenchao and Wu, Feijie},
  journal={Advances in Neural Information Processing Systems},
  volume={34},
  pages={10092--10104},
  year={2021}
}

@article{lin2020ensemble,
  title={Ensemble distillation for robust model fusion in federated learning},
  author={Lin, Tao and Kong, Lingjing and Stich, Sebastian U and Jaggi, Martin},
  journal={Advances in Neural Information Processing Systems},
  volume={33},
  pages={2351--2363},
  year={2020}
}

@article{tan2022fedpcl,
  title={Federated learning from pre-trained models: A contrastive learning approach},
  author={Tan, Yue and Long, Guodong and Ma, Jie and Liu, Lu and Zhou, Tianyi and Jiang, Jing},
  journal={Advances in neural information processing systems},
  volume={35},
  pages={19332--19344},
  year={2022}
}

@inproceedings{li2023bias,
  title={No fear of classifier biases: Neural collapse inspired federated learning with synthetic and fixed classifier},
  author={Li, Zexi and Shang, Xinyi and He, Rui and Lin, Tao and Wu, Chao},
  booktitle={Proceedings of the IEEE/CVF International Conference on Computer Vision},
  pages={5319--5329},
  year={2023}
}

@article{hinton2015distilling,
  title={Distilling the Knowledge in a Neural Network},
  author={Hinton, Geoffrey},
  journal={arXiv preprint arXiv:1503.02531},
  year={2015}
}

@inproceedings{wang2021MulDElogitsKD1,
  title={Mulde: Multi-teacher knowledge distillation for low-dimensional knowledge graph embeddings},
  author={Wang, Kai and Liu, Yu and Ma, Qian and Sheng, Quan Z},
  booktitle={Proceedings of the Web Conference 2021},
  pages={1716--1726},
  year={2021}
}

@inproceedings{yan2020logitsKD2,
  title={Tinygnn: Learning efficient graph neural networks},
  author={Yan, Bencheng and Wang, Chaokun and Guo, Gaoyang and Lou, Yunkai},
  booktitle={Proceedings of the 26th ACM SIGKDD International Conference on Knowledge Discovery \& Data Mining},
  pages={1848--1856},
  year={2020}
}

@inproceedings{yang2021logitsKD3,
  title={Extract the knowledge of graph neural networks and go beyond it: An effective knowledge distillation framework},
  author={Yang, Cheng and Liu, Jiawei and Shi, Chuan},
  booktitle={Proceedings of the web conference 2021},
  pages={1227--1237},
  year={2021}
}

@inproceedings{yang2020structKD1,
  title={Distilling knowledge from graph convolutional networks},
  author={Yang, Yiding and Qiu, Jiayan and Song, Mingli and Tao, Dacheng and Wang, Xinchao},
  booktitle={Proceedings of the IEEE/CVF conference on computer vision and pattern recognition},
  pages={7074--7083},
  year={2020}
}

@inproceedings{guo2022structKD2,
  title={Alignahead: online cross-layer knowledge extraction on graph neural networks},
  author={Guo, Jiongyu and Chen, Defang and Wang, Can},
  booktitle={2022 International Joint Conference on Neural Networks (IJCNN)},
  pages={1--8},
  year={2022},
  organization={IEEE}
}

@inproceedings{huo2023embeddingKD1,
  title={T2-gnn: Graph neural networks for graphs with incomplete features and structure via teacher-student distillation},
  author={Huo, Cuiying and Jin, Di and Li, Yawen and He, Dongxiao and Yang, Yu-Bin and Wu, Lingfei},
  booktitle={Proceedings of the AAAI Conference on Artificial Intelligence},
  volume={37},
  number={4},
  pages={4339--4346},
  year={2023}
}

@inproceedings{he2022embeddingKD2,
  title={Compressing deep graph neural networks via adversarial knowledge distillation},
  author={He, Huarui and Wang, Jie and Zhang, Zhanqiu and Wu, Feng},
  booktitle={Proceedings of the 28th ACM SIGKDD conference on knowledge discovery and data mining},
  pages={534--544},
  year={2022}
}

@inproceedings{yu2022embeddingKD3,
  title={Sail: Self-augmented graph contrastive learning},
  author={Yu, Lu and Pei, Shichao and Ding, Lizhong and Zhou, Jun and Li, Longfei and Zhang, Chuxu and Zhang, Xiangliang},
  booktitle={Proceedings of the AAAI Conference on Artificial Intelligence},
  volume={36},
  number={8},
  pages={8927--8935},
  year={2022}
}

@article{zhao2022graphAug,
  title={Graph data augmentation for graph machine learning: A survey},
  author={Zhao, Tong and Jin, Wei and Liu, Yozen and Wang, Yingheng and Liu, Gang and G{\"u}nnemann, Stephan and Shah, Neil and Jiang, Meng},
  journal={arXiv preprint arXiv:2202.08871},
  year={2022}
}

@inproceedings{2017FedAvg,
  title={Communication-efficient learning of deep networks from decentralized data},
  author={McMahan, Brendan and Moore, Eider and Ramage, Daniel and Hampson, Seth and y Arcas, Blaise Aguera},
  booktitle={Artificial Intelligence and Statistics},
  pages={1273--1282},
  year={2017}
}

@InProceedings{jknet,
  title = 	 {Representation Learning on Graphs with Jumping Knowledge Networks},
  author =       {Xu, Keyulu and Li, Chengtao and Tian, Yonglong and Sonobe, Tomohiro and Kawarabayashi, Ken-ichi and Jegelka, Stefanie},
  booktitle = 	 {Proceedings of the 35th International Conference on Machine Learning},
  pages = 	 {5453--5462},
  year = 	 {2018},
  editor = 	 {Dy, Jennifer and Krause, Andreas},
  volume = 	 {80},
  series = 	 {Proceedings of Machine Learning Research},
  month = 	 {10--15 Jul},
  publisher =    {PMLR}
}

@article{Louvain,
doi = {10.1088/1742-5468/2008/10/P10008},
url = {https://dx.doi.org/10.1088/1742-5468/2008/10/P10008},
year = {2008},
month = {oct},
publisher = {},
volume = {2008},
number = {10},
pages = {P10008},
author = {Vincent D Blondel and Jean-Loup Guillaume and Renaud Lambiotte and Etienne Lefebvre},
title = {Fast unfolding of communities in large networks},
journal = {Journal of Statistical Mechanics: Theory and Experiment}
}

@inproceedings{LargeCK,
author = {Ji, Houye and Zhu, Junxiong and Shi, Chuan and Wang, Xiao and Wang, Bai and Zhang, Chaoyu and Zhu, Zixuan and Zhang, Feng and Li, Yanghua},
title = {Large-scale Comb-K Recommendation},
year = {2021},
isbn = {9781450383127},
publisher = {Association for Computing Machinery},
address = {New York, NY, USA},
url = {https://doi.org/10.1145/3442381.3449924},
doi = {10.1145/3442381.3449924},
booktitle = {Proceedings of the Web Conference 2021},
pages = {2512–2523},
numpages = {12},
location = {Ljubljana, Slovenia},
series = {WWW '21}
}

@inproceedings{WebScale,
author = {Ying, Rex and He, Ruining and Chen, Kaifeng and Eksombatchai, Pong and Hamilton, William L. and Leskovec, Jure},
title = {Graph Convolutional Neural Networks for Web-Scale Recommender Systems},
year = {2018},
isbn = {9781450355520},
publisher = {Association for Computing Machinery},
address = {New York, NY, USA},
url = {https://doi.org/10.1145/3219819.3219890},
doi = {10.1145/3219819.3219890},
booktitle = {Proceedings of the 24th ACM SIGKDD International Conference on Knowledge Discovery \& Data Mining},
pages = {974–983},
numpages = {10},
location = {London, United Kingdom},
series = {KDD '18}
}

@article{yi2023fedssa,
  title={FedSSA: Semantic similarity-based aggregation for efficient model-heterogeneous personalized federated learning},
  author={Yi, Liping and Yu, Han and Shi, Zhuan and Wang, Gang and Liu, Xiaoguang and Cui, Lizhen and Li, Xiaoxiao},
  journal={arXiv preprint arXiv:2312.09006},
  year={2023}
}

@article{fang2024noise,
  title={Noise-robust federated learning with model heterogeneous clients},
  author={Fang, Xiuwen and Ye, Mang},
  journal={IEEE Transactions on Mobile Computing},
  year={2024},
  publisher={IEEE}
}

@article{yang2023allosteric,
  title={Allosteric feature collaboration for model-heterogeneous federated learning},
  author={Yang, Baoyao and Yuen, Pong C and Zhang, Yiqun and Zeng, An},
  journal={IEEE Transactions on Neural Networks and Learning Systems},
  year={2023},
  publisher={IEEE}
}

@article{zhang2024improving,
  title={Improving generalization and personalization in model-heterogeneous federated learning},
  author={Zhang, Xiongtao and Wang, Ji and Bao, Weidong and Zhang, Yaohong and Zhu, Xiaomin and Peng, Hao and Zhao, Xiang},
  journal={IEEE Transactions on Neural Networks and Learning Systems},
  year={2024},
  publisher={IEEE}
}

@article{wu2024fiarse,
  title={FIARSE: Model-heterogeneous federated learning via importance-aware submodel extraction},
  author={Wu, Feijie and Wang, Xingchen and Wang, Yaqing and Liu, Tianci and Su, Lu and Gao, Jing},
  journal={Advances in Neural Information Processing Systems},
  volume={37},
  pages={115615--115651},
  year={2024}
}

@article{alam2022fedrolex,
  title={Fedrolex: Model-heterogeneous federated learning with rolling sub-model extraction},
  author={Alam, Samiul and Liu, Luyang and Yan, Ming and Zhang, Mi},
  journal={Advances in neural information processing systems},
  volume={35},
  pages={29677--29690},
  year={2022}
}

@article{lu2021heterogeneous,
  title={Heterogeneous model fusion federated learning mechanism based on model mapping},
  author={Lu, Xiaofeng and Liao, Yuying and Liu, Chao and Lio, Pietro and Hui, Pan},
  journal={IEEE Internet of Things Journal},
  volume={9},
  number={8},
  pages={6058--6068},
  year={2021},
  publisher={IEEE}
}

@article{liu2024model,
  title={Model-Heterogeneous Federated Graph Learning with Prototype Propagation},
  author={Liu, Zhi and Zhou, Hanlin and He, Xiaohua and Yuan, Haopeng and Du, Jiaxin and Wang, Mengmeng and Shen, Guojiang and Xia, Feng and others},
  journal={IEEE Transactions on Artificial Intelligence},
  year={2024},
  publisher={IEEE}
}
\balance
\clearpage

\appendix
\noindent
{\large\textbf{Outline}}

To meet page limits, we have included as much essential material as possible in the main text. The appendix provides supplementary details that complete our study. It is organized as follows: 
\begin{description}
    \item[~\ref{app:complexity}] Algorithm Complexity Analysis of existing FGL studies.
    \item[~\ref{app: proof}] The Proof of Theorem Analysis presented in Sec.~\ref{sec: Theoretical Analysis}.
    \item[~\ref{app: Detailed Implementation of Experiment}]  Detailed Implementations of Experiment.
\end{description}

\begin{table*}[t]
\caption{Algorithm complexity analysis for existing prevalent FL and FGL studies.}
\vspace{-3mm}
\label{tab: algorithm_analysis}
\resizebox{\linewidth}{!}{
\begin{tabular}{l|ccc|ccc}
\Xhline{1pt}
Method & Client Memory & Server Memory & Inference Memory & Client Time & Server Time & Inference Time \\
\hline
FedAvg & $\mathcal{O}((b+k)f+f^2)$ & $\mathcal{O}(Nf^2)$ & $\mathcal{O}((b+k)f+f^2)$ & $\mathcal{O}(kmf+nf^2)$ & $\mathcal{O}(N)$ & $\mathcal{O}(kmf+nf^2)$ \\
MOON & $\mathcal{O}((b+k)f+Qf^2)$ & $\mathcal{O}(Nf^2)$ & $\mathcal{O}((b+k)f+f^2)$ & $\mathcal{O}(kmf+nf^2+Qnf)$ & $\mathcal{O}(N)$ & $\mathcal{O}(kmf+nf^2)$ \\
FedSage+ & $\mathcal{O}(L((n+sg)f+f^2))$ & $\mathcal{O}(LtNf^2)$ & $\mathcal{O}(L(n+sg)f+Lf^2)$ & $\mathcal{O}(L((m+sg)f+(n+sg)f^2))$ & $\mathcal{O}(N)$ & $\mathcal{O}(L((m+sg)f+(n+sg)f^2))$ \\
FGSSL & $\mathcal{O}(Q((b+k)f+f^2))$ & $\mathcal{O}(Nf^2)$ & $\mathcal{O}(L(b+k)f+Lf^2)$ & $\mathcal{O}(Qkmf+Qnf^2)$ & $\mathcal{O}(N)$ & $\mathcal{O}(kmf+nf^2)$ \\
Fed-PUB & $\mathcal{O}(M((b+k)f+f^2)+M^2)$ & $\mathcal{O}(N(f^2 +M) + P_g)$ & $\mathcal{O}(M(b+k)f+Mf^2)$ & $\mathcal{O}(Mkmf+Mnf^2)$ & $\mathcal{O}(N^2(\log(N)+M^2))$ & $\mathcal{O}(Mkmf+Mnf^2)$ \\
FedGTA & $\mathcal{O}((b+k)f+f^2+kKc)$ & $\mathcal{O}(Nf^2+NkKc)$ & $\mathcal{O}((b+k)f+f^2)$ & $\mathcal{O}(km(f+knc)+n(f^2+c))$ & $\mathcal{O}(N+NkKc)$ & $\mathcal{O}(kmf+nf^2)$ \\
AdaFGL & $\mathcal{O}(E((b+k)f+f^2))$ & $\mathcal{O}(Nf^2)$ & $\mathcal{O}(E((b+k)f+f^2))$ & $\mathcal{O}(E(kmf+nf^2))$ & $\mathcal{O}(N)$ & $\mathcal{O}(E(kmf+nf^2))$ \\
\hline
FedGKC & $\mathcal{O}(2(b+l)f+2f^2)$ & $\mathcal{O}(Nf^2)$ & $\mathcal{O}(2(b+l)f+2f^2)$ & $\mathcal{O}(kmf+nf^2+f^2)$ & $\mathcal{O}(N)$ & $\mathcal{O}(kmf+nf^2)$ \\
\Xhline{1pt}
\end{tabular}}
\end{table*}

\section{Algorithm Complexity Analysis}
\label{app:complexity}
    Table~\ref{tab: algorithm_analysis} summarizes the time and space complexity of different methods.
    Here, $n$, $m$, $c$, and $f$ denote the numbers of nodes, edges, classes, and feature dimensions, respectively.
    $s$ refers to the number of selected perturbated nodes, and $g$ indicates the number of generated neighbors.
    $b$ is the batch size.
    Parameters $k$ and $K$ represent the number of feature aggregation steps and the order of moments, respectively. 
    $N$ denotes the number of participating clients per round. $Q$ corresponds to the query set size used in contrastive learning.
    $E$ stands for the number of models in ensemble learning, $M$ is the dimension of the trainable matrix for masking weights. $P_g$ denotes the size of the pseudo-graph stored on the server.

    Local updates play a key role in federated training efficiency.
    For example, for MOON and FGSSL, the cost of contrastive learning depends on the size and semantics of the query set, resulting in complexities of $\mathcal{O}(Qf^2)$ and $\mathcal{O}(Q((b+k)f+f^2))$, respectively, which may become prohibitive as local data grows.
    Ensemble-based AdaFGL maintains multiple local models to capture private semantics, bounded by $\mathcal{O}(E((b+k)f+f^2))$.
    Some approaches rely on additional message exchange.
    Specifically, FedSage+ requires client-to-client message sharing for subgraph augmentation, with complexity $\mathcal{O}(L((n+sg)f+f^2))$.
    Fed-PUB maintains a global pseudo-graph on the server and leverages uploaded weights for personalization, introducing a complexity of $\mathcal{O}(N(f^2+M)+P_g)$.
    In contrast, FedGTA is relatively lightweight, using topology-aware soft labels for personalized aggregation on the server, but the extra encoding signal still incurs a complexity of $\mathcal{O}(kKC)$.

    Compared with the above methods, our FedGKC demonstrates superior scalability in both time and space complexity.
    Its overall cost can be approximated as $\mathcal{O}(N((b+l)f+f^2))$, which grows linearly with the number of clients and standard graph operations, without relying on large query sets, global pseudo-graphs, or multiple locally stored models.
    Moreover, FedGKC leverages lightweight self- and mutual-distillation for cross-model knowledge alignment on clients, and adopts a knowledge-aware aggregation scheme on the server, thereby avoiding heavy communication or redundant parameter updates.
    This enables FedGKC to maintain low computational and storage overhead while significantly improving accuracy, striking a favorable balance between efficiency and performance, which  highlight FedGKC's advantage over existing baselines.

\section{The Proof of Theoretical Analysis}
\label{app: proof}
    In this section, we present comprehensive derivations and rigorous proofs for Theorem~\ref{Node Score Expression} and Theorem~\ref{Inner Product differences}.

\subsection{The Proof of Theorem~\ref{Node Score Expression}}
    This theorem establishes the relationship between the node score $S_i$ and its underlying components of knowledge strength and knowledge clarity, thereby characterizing how the scoring mechanism reflects both confidence and smoothness.

\paragraph{Proof.} 
    Because $\sum_{c} p_{i,c} = 1$ and $\sum_{c \neq \arg\max p_i} p_{i,c} = 1 - m_i$, we can reformulate the definition of knowledge intensity into the following mathematical expression:
\begin{equation}
    Q_i^{\text{clar}} = \frac{1}{M-1} \left( m_i - (1 - m_i) \right) - \lambda \,\overline{\text{sim}}_i= \frac{2m_i - 1}{M-1} - \lambda \,\overline{\text{sim}}_i.
\end{equation}
    Accordingly, the node score $S_i$ is formulated as the aggregation of knowledge strength and knowledge clarity, which can be expressed as Eq.~(\ref{Node Score}).
    Subsequently, we compute the partial derivatives of $m_i$ and $\overline{\mathrm{sim}}_i$ in Eq.~(\ref{Node Score}), which leads to the following result:
\begin{equation}
    \frac{\partial S_i}{\partial m_i} = \alpha = 1 + \frac{2}{M-1}, \;\;\; \frac{\partial S_i}{\partial \overline{\text{sim}}_i} = -\lambda.
\end{equation}
    The first partial derivative is strictly positive, whereas the second partial derivative is strictly negative.
    
    In summary, this proof shows that the proposed aggregation weights assign higher scores to nodes with greater confidence, while simultaneously penalizing excessive oversmoothing.
    This property ensures that the scoring mechanism captures meaningful knowledge quality in a theoretically consistent manner.

\subsection{The Proof of Theorem~\ref{Inner Product differences}}
    This theorem formally characterizes the discrepancy between the alignment of KAMA and FedAvg, and proves that the gradient direction induced by KAMA approaches the global optimal gradient direction more closely.

\paragraph{Proof.}
    According to the definition of the knowledge-related weight, we obtain:
\begin{equation}
    \sum_{k} w_k^{\text{know}} a_k = \frac{\sum_{k} P_k a_k}{\sum_{j} P_j} = \frac{K \left( \overline{P}\,\overline{a} + \mathrm{Cov}(P,a) \right)}{K \overline{P}} = \overline{a} + \frac{\mathrm{Cov}(P,a)}{\overline{P}}.
\end{equation}

    Initially, we need to unify the numerator and denominator.
\begin{equation}
    \sum_{k} w_k^{\text{know}} a_k= \frac{\sum_{k} P_k a_k}{\sum_{j} P_j}= \frac{\sum_{k} P_k a_k}{K\,\overline{P}}.
\label{eq:weighted-average-start}
\end{equation}

    Next, we decompose the numerator by writing $P_k = \left(P_k-\overline{P}\right)+\overline{P}$:
\begin{equation}
\begin{aligned}
    \sum_{k} P_k a_k&= \sum_{k} \left(\left(P_k-\overline{P}\right)+\overline{P}\right) a_k= \overline{P}\sum_{k} a_k + \sum_{k} \left(P_k-\overline{P}\right)a_k \\
    &= K\overline{P}\,\overline{a}+ \sum_{k} \left(P_k-\overline{P}\right)\left(\left(a_k-\overline{a}\right)+\overline{a}\right) \\
    &= K\overline{P}\overline{a}+ \underbrace{\sum_{k} \left(P_k-\overline{P}\right)\left(a_k-\overline{a}\right)}_{\displaystyle K\mathrm{Cov}\left(P,a\right)}+ \overline{a}\underbrace{\sum_{k}\left(P_k-\overline{P}\right)}_{\displaystyle 0} \\
    &= K\overline{P}\overline{a} + K\mathrm{Cov}\left(P,a\right).
\label{eq:numerator-decompose}
\end{aligned}
\end{equation}
    Here we use $\sum_{k}\left(P_k-\overline{P}\right)=0$ and the covariance definition:
\begin{equation}
    \mathrm{Cov}(P,a) := \frac{1}{K}\sum_{k=1}^{K} \left(P_k-\overline{P}\right)\left(a_k-\overline{a}\right).
\end{equation}
    Substituting this decomposition back into the previous expression, we obtain:
\begin{equation}
    \sum_{k} w_k^{\text{know}} a_k= \frac{K\left(\overline{P}\,\overline{a}+\mathrm{Cov}\left(P,a\right)\right)}{K\,\overline{P}}= \overline{a} + \frac{\mathrm{Cov}\left(P,a\right)}{\overline{P}}.
\end{equation}

    Finally, this leads to the formulation of KAMA, whose inner product can be expressed as:
\begin{equation}
\begin{aligned}
    \langle \hat{g}_{\text{KAMA}}, g^\star \rangle&= \sum_{k} \frac{1}{2}\left(w_k^{\text{vol}} + w_k^{\text{know}}\right)a_k \\
    &= \frac{1}{2}\sum_{k} w_k^{\text{vol}} a_k + \frac{1}{2}\sum_{k} w_k^{\text{know}} a_k \\
    &= \frac{1}{2}\underbrace{\langle \hat{g}_{\text{FedAvg}}, g^\star \rangle}_{\sum_{k} w_k^{\text{vol}} a_k}+ \frac{1}{2}\left(\overline{a} + \frac{\mathrm{Cov}(P,a)}{\overline{P}}\right).
\end{aligned}
\end{equation}

    Consequently, Eq.~(\ref{Differences Equation}) is established.
    This positive discrepancy confirms that the gradient descent direction under KAMA exhibits superior alignment with the global optimum compared to FedAvg.

\section{Detailed Implementation of Experiment}
\label{app: Detailed Implementation of Experiment}

Table~\ref{tab:dataset} display the statistical information of selected datasets for our experiment, which have been widely applied in Graph Machine Learning research. 
This section displays dataset descriptions and experimental environments. 

\begin{table}[t]
\caption{Statistics of the eight employed benchmark datasets.}
\vspace{-3mm}
\label{tab:dataset}
\resizebox{\linewidth}{!}{
\begin{tabular}{llllcc}
\Xhline{1pt}
Dataset & Nodes & Features & Edges & Classes & Train/Val/Test \\
\hline
Cora & 2,708 & 1,433 & 5,429 & 7 &20\%/40\%/40\% \\
CiteSeer & 3,327 & 3,703 & 4,732 & 6 &20\%/40\%/40\% \\
PubMed & 19,717 & 500 & 44,338 & 3 &20\%/40\%/40\% \\
CS  & 18,333 & 6,805 & 81,894 & 15 &20\%/40\%/40\% \\
Physics & 34,493 & 8,415 & 247,692 & 5 &20\%/40\%/40\% \\
Photo & 7,487 & 745 & 119,043 & 8 &20\%/40\%/40\% \\
Computer & 13,381 & 767 & 245,778 & 10 &20\%/40\%/40\% \\
ogbn-arxiv & 169,343 & 128 & 2,315,598 & 40 &60\%/20\%/20\% \\
\Xhline{1pt}
\end{tabular}}
\end{table}
\subsection{Toy Experiment Implementation}
\label{appendix:toy_example}
The simulation settings and experimental environment is aligned with description in Sec.~\ref{sec: Experimental Setup}. 
Experiment is conducted on the Cora dataset, which is further partitioned into 10 clients.
Representative baselines selected for each category are most competitive MHtFGL methods to our knowledge: FML~\cite{shen2023fedfml} for Partial Model, FedPPN~\cite{liu2024model} for Prototype, and AlFeCo~\cite{yang2023allosteric} for Public/Synthetic Information.  

\subsection{Dataset Descriptions}
\textbf{Cora}, \textbf{CiteSeer}, and \textbf{PubMed} are three citation network datasets. Nodes represent papers and edges represent citation relationships. 
    The node features are word vectors, where each element indicates the presence or absence of each word. 

\noindent
\textbf{CS}, \textbf{Physics} are co-authorship graphs derived from the MAG. 
    In these graphs, nodes represent individual authors, edges denote co-authorship relationships between them, and node features are constructed from the keywords of the authors' publications. 
    The labels assigned to the nodes indicate the specific research fields in which the authors are active. 
    These datasets are commonly used for evaluating graph-based methods, particularly for node classification.

\noindent\textbf{Photo} and \textbf{Computers} are segments of the Amazon co-purchase graph, where nodes represent items and edges represent that two goods are frequently bought together. 
    Given product reviews as bag-of-words node features.
    
\noindent
\textbf{ogbn-arxiv} are two citation graphs indexed by MAG.
    Each paper involves averaging the embeddings of words in its title and abstract.
    The embeddings of individual words are computed by running the skip-gram model.
    
\subsection{Implementation Details} 
    The experiments are conducted on an NVIDIA GeForce RTX 3090 GPU.
    Unless otherwise specified, the copilot model employed is a two-layer GCN architecture.
    The evaluation metric focuses on node classification accuracy within subgraphs on the client side, with performance averaged across all clients.
    The Adam optimizer is selected for the optimization process.
    For hyperparameter values, $\alpha$ and $\beta$ in Eq.~(\ref{equ:copmodel}) and Eq.~(\ref{equ:localmutu}) are set to 0.6 and 0.2, respectively, and $\lambda$ in Eq.~(\ref{equ:clarity}) is set to 0.1.
    The evaluation metric is the node classification accuracy on the test set.
    To ensure reproducibility, the complete code for implementing the FedGKC is released publicly at \url{https://anonymous.4open.science/r/FedGKC-65BD/}.

\end{document}